\documentclass{SCIS2021}

\usepackage[english]{babel}
\usepackage{amsthm}
\usepackage{bm}
\usepackage[colorlinks, linkcolor=black]{hyperref}
\newtheoremstyle{exampstyle}{0.0em} 
{0.0em} 
{} 
{1em} 
{\bfseries} 
{.} 
{1em} 
{} 
\usepackage{balance}
\usepackage{multirow}  
\theoremstyle{exampstyle}

\usepackage{CJKutf8}
\usepackage{graphicx}
\usepackage{colortbl} 
\usepackage[table]{xcolor}
\usepackage[flushleft]{threeparttable}
\usepackage{pifont}
\usepackage{tcolorbox}  
\newcommand{\cmark}{\ding{51}}%
\newcommand{\xmark}{\ding{55}}%
\usepackage{arydshln} 

\makeatletter

\definecolor{mycolor}{RGB}{235, 235, 235} 
\definecolor{baise}{RGB}{255,255,255}

\makeatother
\begin{document}
	\ArticleType{RESEARCH PAPER}
	\Year{2025} \Month{1} \Vol{} \No{} \DOI{} \ArtNo{} \ReceiveDate{} \ReviseDate{}
	\AcceptDate{} \OnlineDate{}
		
            \AuthorMark{Shenxi L}
            \author[1]{Shenxi Liu}{}  
            \author[1]{Kan Li}{}  
            \author[2]{Mingyang Zhao}{}  
            \author[1]{Yuhang Tian}{}  
            \author[3]{Bin Li}{{b.li2@siat.ac.cn}}  %
            \author[3]{Shoujun Zhou}{} 
            \author[4]{\\Hongliang Li}{}
            \author[5]{Fuxia Yang}{}

            \AuthorCitation{Shenxi Liu, Kan Li, Mingyang Zhao, Yuhang Tian, Bin Li, Shoujun Zhou, Hongliang Li, Fuxia Yang}

            
            \address[1]{Beijing Institute of Technology, Beijing {\rm 100081}, China}   
            \address[2]{Department of Computing, The Hong Kong Polytechnic University, 999077, Hong Kong}
            \address[3]{Shenzhen Institute of Advanced Technology, Chinese Academy of Sciences, Shenzhen, {\rm 518067}, China}   
            \address[4]{Dept. of Med. Engineering of Air Force Hospital of Southern Theater Command of PLA, Guangdong {\rm 510180}, China}
            \address[5]{Acupuncture Department, Shenzhen Traditional Chinese Medicine Hospital, Shenzhen {\rm 518033}, China}
            
		\title{$\text{M}^3\text{-Med}$: A Benchmark for Multi-lingual, Multi-modal, and Multi-hop Reasoning in Medical Instructional Video Understanding}{$\text{M}^3\text{-Med}$: A Benchmark for Multi-modal, Multi-lingual, and Multi-hop Reasoning in Medical Instructional Video Understanding}
\abstract{
With the rapid progress of artificial intelligence (AI) in multi-modal understanding, there is increasing potential for video comprehension technologies to support professional domains such as medical education.
However, existing benchmarks suffer from two primary limitations: (1) Linguistic Singularity: they are largely confined to English, neglecting the need for multilingual resources; and (2) Shallow Reasoning: their questions are often designed for surface-level information retrieval, failing to properly assess deep multi-modal integration.
To address these limitations, we present M$^3$-Med, the first benchmark for Multi-lingual, Multi-modal, and Multi-hop reasoning in Medical instructional video understanding. 
M$^3$-Med consists of medical questions paired with corresponding video segments, annotated by a team of medical experts. 
A key innovation of M$^3$-Med is its multi-hop reasoning task, which requires a model to first locate a key entity in the text, then find corresponding visual evidence in the video, and finally synthesize information across both modalities to derive the answer.
This design moves beyond simple text matching and poses a substantial challenge to a model's deep cross-modal understanding capabilities.
We define two tasks: Temporal Answer Grounding in Single Video (TAGSV) and Temporal Answer Grounding in Video Corpus (TAGVC). 
We evaluated several state-of-the-art models and Large Language Models (LLMs) on M$^3$-Med. The results reveal a significant performance gap between all models and human challengers, especially on the complex multi-hop questions where model performance drops sharply.
M$^3$-Med effectively highlights the current limitations of AI models in deep cross-modal reasoning within specialized domains and provides a new direction for future research.
}


		\maketitle
            \section{Introduction}
AI technology is profoundly reshaping our interaction with digital information, and a central challenge in this field is enabling machines to achieve a human-level understanding of complex video content \cite{li2024towards, wu2025generative, AVSGN}.
In high-value domains such as medical education, clinical training, and patient communication, the intelligent analysis of video materials offers immense potential \cite{zhang2023multi,TemporalWM,Video-MM}. This technology can help medical students quickly locate key knowledge points, assist doctors in case reviews, and even intuitively explain complex medical procedures to patients \cite{Blackbox}.

Video Question Answering (VideoQA) and Temporal Video Grounding have become standard tasks for evaluating a model's video comprehension \cite{wen2024learning, zong2025ask2loc}. 
In the open domain, benchmarks like ActivityNet \cite{ActivityNet} and QVHighlights \cite{QVHighlights} have driven significant progress. Subsequently, research has expanded into specialized domains, with datasets such as YouCook2 \cite{YouCook2,YouCook2-ex} for cooking and MovieNet \cite{MovieNet} for films, demonstrating the value of domain-specific video understanding.
Following this trend, several benchmarks have been introduced for the medical domain, including NurVid \cite{NurVid}, MedVid \cite{MedVid} and HealthVid \cite{HealthVid}.
However, despite these advances, existing benchmarks, particularly in medicine, suffer from critical limitations.

These benchmarks suffer from two primary limitations. First, they are predominantly monolingual (English), failing to address the global need for multilingual medical information access. Second, they are susceptible to 'shortcut learning,' where questions can often be answered by relying solely on subtitles through simple keyword matching, without requiring genuine comprehension of the visual content (see Fig. \ref{SimpleQuestion}).
This reliance on shortcuts prevents a genuine evaluation of a model's ability to deeply integrate and reason over information from different modalities \cite{wu2025image,CrackNex,MMKGR,Shallow,Li2024}.

\begin{figure}[h]
    \input{Figures/6.2.Reasoning}
\end{figure}

To address these challenges, we propose \textbf{M$^3$-Med}\footnote{The benchmark is publicly available at \href{https://Med-M3-Dataset.github.io/}{\textcolor{blue}{\underline{{https://Med-M3-Dataset.github.io/}}}}.}, the first benchmark designed for \textbf{M}ulti-lingual, \textbf{M}ulti-modal, and \textbf{M}ulti-hop reasoning on \textbf{Med}ical instructional videos.

Our dataset is constructed from instructional videos sourced from public platforms and annotated by a team of medical experts. 
A key innovation of M$^3$-Med is its two-tier question design. 
``Simple questions" can be answered via direct information retrieval, similar to traditional benchmarks. 
In contrast, ``complex questions" necessitate multi-hop reasoning: they require the model to first identify a key concept in the text, then locate corresponding visual evidence in the video, and finally synthesize these cross-modal cues to derive the correct answer.

We benchmarked a range of state-of-the-art models, including Large Language Models (LLMs) and Multi-modal LLMs (MLLMs), against human challengers performance. 
Our experiments show that while human challengers achieve near-perfect accuracy, all models perform significantly worse, with their performance dropping sharply on complex, multi-hop questions. 
These results demonstrate that M$^3$-Med effectively exposes the limitations of current models in deep, cross-modal reasoning.

Our main contributions are threefold: 
(1) We construct and release M$^3$-Med, a new high-quality dataset for evaluating multi-hop, cross-modal reasoning in the medical domain. 
(2) We propose a novel two-tier question design paradigm that moves beyond shallow text matching to assess a model's true reasoning depth. 
(3) We provide extensive benchmarks with state-of-the-art models, revealing their current capabilities and outlining directions for future research.

This work is a continuation of the NLPCC 2023 Shared Task 5: Chinese Medical Instructional Video Question Answering \footnote{Visit \href{https://github.com/cmivqa/NLPCC-2023-Shared-Task-5}{\textcolor{blue}{\underline{{https://github.com/cmivqa/NLPCC-2023-Shared-Task-5}}}} for details.} \cite{NLPCC2023} and the NLPCC 2024 Shared Task 7: Multilingual Medical Instructional Video Question Answering Challenge \cite{NLPCC2024} \footnote{\href{https://github.com/Lireanstar/NLPCC2024_MMIVQA}{\textcolor{blue}{\underline{{https://github.com/Lireanstar/NLPCC2024\_MMIVQA}}}}}. 
Furthermore, the dataset from our work served as the basis for the NLPCC 2025 Shared Task 4: Multi-modal, Multi-lingual, and Multi-hop Medical Instructional Video Question Answering Challenge \cite{NLPCC2025} \footnote{\href{https://github.com/cmivqa/NLPCC2025_M4IVQA}{\textcolor{blue}{\underline{{https://github.com/cmivqa/NLPCC2025\_M4IVQA}}}}}.

            \section{Related Work}
The task of localizing specific video segments based on natural language queries, known as Video Temporal Grounding (VTG), has become a key area of research. 
Driven by the proliferation of online video and advances in Visual Language Models (VLMs), VTG has seen growing demand in applied domains like education and medicine, leading to the proposal of numerous benchmarks and methods.

\subsection{Benchmarks for Medical Video Understanding}
Several benchmarks have been established for medical video understanding.
For instance, NurVid \cite{NurVid} provides a dataset focused on nursing education, with tasks centered on classifying procedures and detecting specific nursing actions within videos.

Similarly, the work by \cite{MedVid} introduced two datasets: MedVidCL, designed for classifying videos into medical-instructional, medical-non-instructional, and non-medical categories; and MedVidQA, which focuses on the temporal localization of answers to natural language questions in medical videos.

Building on MedVidQA, \cite{HealthVid} proposed a pipeline to automatically generate larger-scale datasets, resulting in HealthVidQA-CRF and HealthVidQA-Prompt, which significantly increased the volume of available data for this task.

We summarize and compare these existing medical video datasets with our work in Tab. \ref{Comp}. As the table highlights, all prior benchmarks are limited to English and, critically, are designed for single-step retrieval rather than multi-hop reasoning.

\begin{table}[h]  
        \centering  
    \caption{
    A comparison of existing medical video understanding datasets.
    }  
    \resizebox{1.0\textwidth}{!}{
        \begin{tabular}{lcccccc}  
            \toprule[2pt] 
            $\begin{array}{l}\textbf{Dataset}\end{array}$	&	\textbf{Multi-Hop}	&	\textbf{Language(s)}	&	\textbf{Annotation Method}	&	\textbf{Task}	&	\textbf{Number of Videos}	&	\textbf{Number of Questions}	\\
            \midrule													
            $\begin{array}{l}\text{Nurvid} \end{array}$ \cite{NurVid}	&	\textcolor{red}{\xmark}	&	English	&	Manually	&	Choose. and Detect.	&	$\begin{array}{c}\text{1,538} \end{array}$	&	$\backslash$	\\
            \midrule													
            $\begin{array}{l}\text{MedVidCL} \end{array}$ \cite{MedVid}	&	\textcolor{red}{\xmark}	&	English	&	Manually	&	Choose.	&	$\begin{array}{c}\text{1,016 Medical} \\  \text{(3,842 Non-Medical)} \end{array}$	&	$\backslash$	\\
            \midrule													
            $\begin{array}{l}\text{MedVidQA} \end{array}$\cite{MedVid}	&	\textcolor{red}{\xmark}	&	English	&	Manually	&	Local.	&	$\begin{array}{c}\text{899} \end{array}$	&	$\begin{array}{c}\text{3,010} \end{array}$	\\
            \midrule													
            $\begin{array}{l}\text{HealthVidQA-CRF}\end{array}$ \cite{HealthVid}	&	\textcolor{red}{\xmark}	&	English	&	Automatic$^\dagger$	&	Local.	&	$\begin{array}{c}\text{11,708} \end{array}$	&	$\begin{array}{c}\text{23,434} \end{array}$	\\
            \midrule													
            $\begin{array}{l}\text{HealthVidQA-Prompt} \end{array}$ \cite{HealthVid}	&	\textcolor{red}{\xmark}	&	English	&	Automatic$^\dagger$	&	Local.	&	$\begin{array}{c}\text{13,990} \end{array}$	&	$\begin{array}{c}\text{52,771} \end{array}$	\\
            \midrule													
            $\begin{array}{l}\textbf{ M$^3$-Med }\\ \textbf{ (Ours) }\end{array}$	&	\textcolor{green}{\cmark}	&	Chinese and English	&	Manually	&	Choose. and Local.	&	$\begin{array}{c}\text{3,748} \end{array}$	&	$\begin{array}{c}\text{12,747} \end{array}$	\\
            \bottomrule[2pt]  
        \end{tabular}    
    }
    \begin{tablenotes}
        \footnotesize
        \item[*] Tasks are defined as: Choose. (multiple-choice selection); Detect. (action detection); Local. (temporal localization). $\dagger$ Indicates annotations were generated automatically, relying primarily on textual information.
    \end{tablenotes}
    \label{Comp}

\end{table}

\subsection{Cross Multi-Modal Multi-Hop Reasoning}
While the aforementioned datasets have advanced the field, they are primarily designed for single-hop information retrieval. In contrast, many real-world scenarios require multi-hop reasoning over a combination of information sources.

To address this gap, our work introduces the concept of multi-hop reasoning, prominently featured in textual QA benchmarks like MMKGR \cite{MMKGR}, to the domain of medical video understanding.

This reasoning paradigm has been explored in other multi-modal contexts that typically involve static images and text. 
For example, FCMR \cite{FCMR} requires reasoning over financial reports and charts, while MMQA \cite{MMQA} and MMKG \cite{MMKG} demand joint reasoning over text, tables, and images.

However, a significant gap remains: these multi-hop benchmarks are confined to static modalities (text and images). 
Our work is the first to integrate this complex reasoning paradigm into a dynamic, video-based temporal grounding task.

\subsection{Methods for Video Temporal Grounding}
The development of VTG benchmarks has spurred the creation of various advanced methods. We survey the most relevant ones here, which serve as the primary baselines for evaluation on our proposed M$^3$-Med dataset.

These include MutualSL \cite{MutualSL}, which uses Mutual Knowledge transfer for Span Localization; a Prompt-Based Learning (PBL) approach \cite{PBL}; CCGS \cite{CCGS}, which employs a Cross-modal Contrastive Global-Span framework; and FMALG \cite{FMALG}, a Fine-grained Modality Alignment and Local-Global optimization method. 
These models have all achieved state-of-the-art (SOTA) performance on related leader-boards, making them suitable for testing the challenges posed by M$^3$-Med.

In addition to these specialized VTG models, the recent emergence of powerful Multi-modal Large Language Models (MLLMs), such as Video-ChatGPT \cite{VideoGPT} and Qwen2.5-VL \cite{QwenVL}, presents a new frontier. Evaluating the capabilities of these general-purpose models on our specialized, multi-hop task is therefore also a key focus of our work.

            \begin{figure}[h]
    \input{Figures/3.2.UIs}
\end{figure}

\section{Building up M$^3$-Med Benchmark}
This section details the construction of M$^3$-Med, a new benchmark for medical video understanding. It is specifically designed to assess a model's capabilities in multi-lingual, multi-modal, and, most critically, multi-hop reasoning.
The final benchmark comprises a collection of medical instructional videos, each paired with a set of natural language questions. We also provide professionally annotated knowledge graphs (KGs) and high-quality subtitles in SRT format for each video.

Our annotation process was executed by a three-tiered team to ensure high quality: 
(1) Question Writers: Professional doctors who created questions based on video content. 
(2) Time-stamp markers: Medical students who marked the temporal boundaries of answers. 
(3) Supervisors: Senior experts who oversaw the entire quality control process.

The annotation pipeline includes following stages.

\begin{figure}[htbp!]
    \centering
    \includegraphics[width=\textwidth]{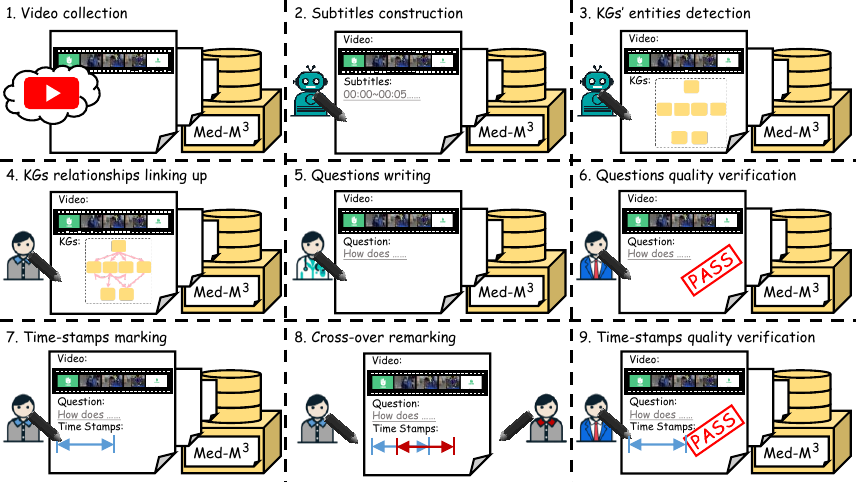}
    \caption{The steps for Annotation and Quality Control. This flowchart details the structured process for data annotation and validation, beginning with initial data creation, followed by multiple rounds of peer-review and expert verification to ensure the highest data quality.} \label{step}
\end{figure}

\subsection{Video Collection}
The quality of our benchmark begins with a careful video selection process.
We adopted a multi-source collection strategy. 
First, following the methodology of \cite{MedVid}, we identified relevant medical and health-related keywords from wikiHow \footnote{\href{https://www.wikihow.com/}{\textcolor{blue}{\underline{{https://www.wikihow.com/}}}}} and used them (and their Chinese translations) to source videos from YouTube \footnote{\href{https://www.youtube.com/}{\textcolor{blue}{\underline{{https://www.youtube.com/}}}}}. 
Second, to broaden our scope, we sampled videos from relevant categories in the HowTo100M dataset \cite{Howto100M}, a strategy adapted from \cite{HealthVid}. 
Finally, every video underwent a manual screening process to verify its relevance to medical instruction and to remove duplicates.

\subsection{Subtitles Construction}
To support models without built-in speech recognition and to enable experiments that test reasoning across text and video, we generated high-quality subtitles for all videos.
We employed the Whisper model \cite{Whisper} to transcribe all videos, producing subtitles in the standard SRT format (see the top-right part of Fig. \ref{KGsConstruct} for an example).

To facilitate subsequent processing, we normalized the subtitles by applying lemmatization to the English text using NLTK \cite{NLTK} and tokenization to the Chinese text using Jieba \cite{jieba}.

\subsection{Knowladge Graphs Construction}
The foundation of our multi-hop reasoning task is a Knowledge Graph (KG) constructed for each video. These KGs are central to our benchmark, providing the explicit entities and relationships required to differentiate between simple (single-hop) and complex (multi-hop) questions.
We employed a human-in-the-loop pipeline, combining automated tools with expert manual curation, to construct the KGs.

\begin{figure}[htbp]
    \centering
    \includegraphics{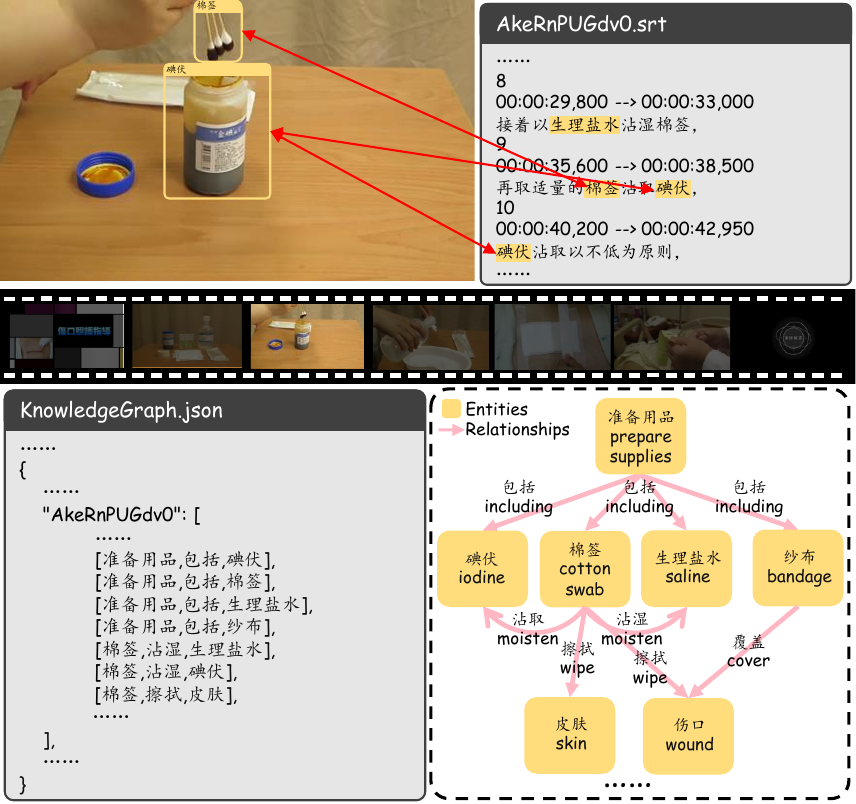}
    \caption{Illustrative Example of Knowledge Graph Construction. This diagram provides a concrete example of how structured knowledge is extracted from video frames (upper-left) and subtitles (upper-right) to construct a knowledge graph (lower-left), showing the identification of entities and their relationships. Its logical structure is shown at the lower-right.} \label{KGsConstruct}
\end{figure}

The construction process, illustrated in Fig. \ref{KGsConstruct}, involved the following steps:
(1) Automated Entity Extraction: We used powerful LLMs (Qwen2.5 \cite{Qwen} for Chinese, GPT-4o \cite{GPT} for English) to extract all potential entities from the subtitles.
(2) Visual Grounding: For each extracted entity, we used Grounding DINO \cite{DINO} to detect its visual presence in the video frames corresponding to the subtitle's timestamp.
(3) Human Verification \& Refinement: Human annotators then reviewed all machine-generated entities, filtering for relevance to the video's medical theme and merging duplicates.
(4) Relation Annotation: Finally, the annotators manually constructed the relationships between the verified entities to complete the KG.

\subsection{Medical Questions Writing}
With the KGs as a foundation, our medical professional team proceeded to write the questions, creating an equal number of ``simple" and ``complex" questions for each video.

To ensure the quality and validity of the questions, a rigorous peer-review process was implemented (Fig. \ref{process2}). 
Each question was checked against three criteria: 
(1) Medical Professionalism: Is the language medically sound?
(2) Content Relevance: Is the question directly related to the video content? 
(3) Label Consistency: Does the question's content accurately reflect its ``simple" or ``complex" reasoning label? 
Disputed questions were revised through team discussion or discarded if they could not meet these standards.

\subsection{Time-Stamps Marking}
Following question creation, annotators identified the precise video segment that answered each question and marked its start and end time-stamps.

To ensure high inter-annotator agreement, a 10\% random sample of the annotations was re-annotated by a different annotator. 
We then quantified the consistency between the original and second annotations using the Cohen’s Kappa ($\kappa$) coefficient.

Assume the sequences marked by the two markers as $T_1$, $T_2$, and the whole video as $T$. Firstly, build up the confusion matrix $C$:

$$C=\begin{bmatrix}a&b \\ c&d\end{bmatrix}$$

where  $a, b, c, d$ represent $|T_1 \cap T_2| , |T\backslash T_1 \cap T_2| , |T_1 \cap T \backslash T_2| , |T\backslash T_1 \cap T \backslash T_2|$. Then, calculate the observed consistency $P_o$ and expected consistency $P_e$ from confusion matrix $C$ as follows:

\begin{align*}
P_o =& \frac{a + d}{N} \\
P_e =& \left( \frac{(a + b)(a + c)}{N^2} \right) + \left( \frac{(c + d)(b + d)}{N^2} \right) 
\end{align*}

Finally, $\kappa$ can be calculated as:

$$\kappa = \frac{P_o - P_e}{1 - P_e} $$
 
Annotations with a $\kappa$ score below 0.25, indicating low agreement, were flagged for manual review and final adjudication by a supervisor.
Fig. \ref{process3} is an example of time-stamps marking and remarking.

\subsection{Tasks Designing}
Inspired by the task proposed by \cite{MedVid,HealthVid}, we designed two tasks for this benchmark.

\textbf{Task 1: Temporal Answer Grounding in Single Video (TAGSV)}

\begin{figure}[H]
    \centering
    \includegraphics{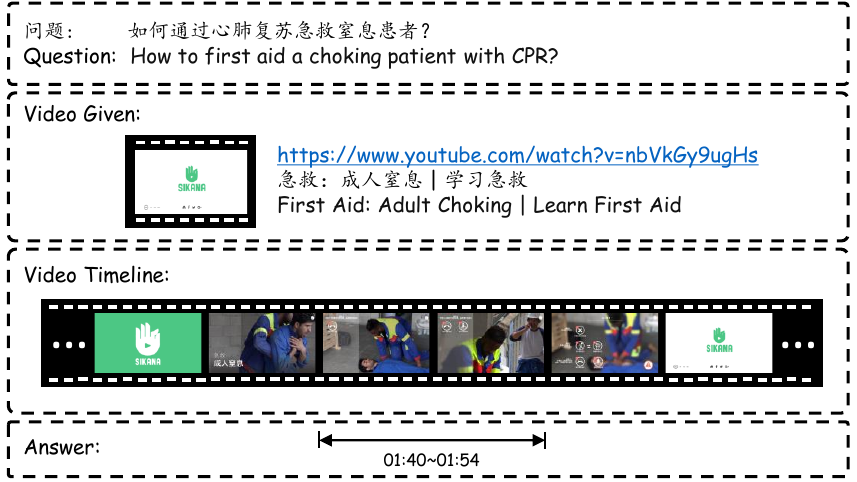}
    \caption{Example of Temporal Answer Grounding in a Single Video (TAGSV). This figure illustrates the TAGSV task, where a model must localize and extract the precise temporal segment from a single video that answers the given natural language query.} \label{Task1}
\end{figure}

Given a medical question and a video, the challenger needs to clip the sequence of the video corresponds to the problem, giving the answer in the form of start time-stamp and end time-stamp. As shown in Fig. \ref{Task1}.

Assume the total number of questions is $n$, and the question $Q_i$ corresponds to the correct answer $T_i^*=[t_{start}^*,t_{end}^*] ,i\in[1,n]$.
The answer to the question given to the question $Q_i$ is $T_i=[t_{start},t_{end}]$.
Let the $IoU$ between the challenger's answer and the correct answer for question $i$ be:

$$\text{IoU}_i=\frac{T_i^* \cap T_i}{T_i^* \cup T_i}$$

The challenger's score was assessed using ``$\mathrm{IoU}_{\tau=0.3/0.5/0.7}$" and ``mIoU", calculated as follows:

$$\text{IoU}_{\tau=c}= \frac{\text{numbers of i which IoU}_i\text{ greater than } c}{n}$$

$$\text{mIoU}=\frac{\sum_{i=1}^n \text{IoU}_i}{n}$$

\textbf{Task 2: Temporal Answer Grounding in Video Corpus (TAGVC)}

\begin{figure}[htbp]
    \centering
    \includegraphics{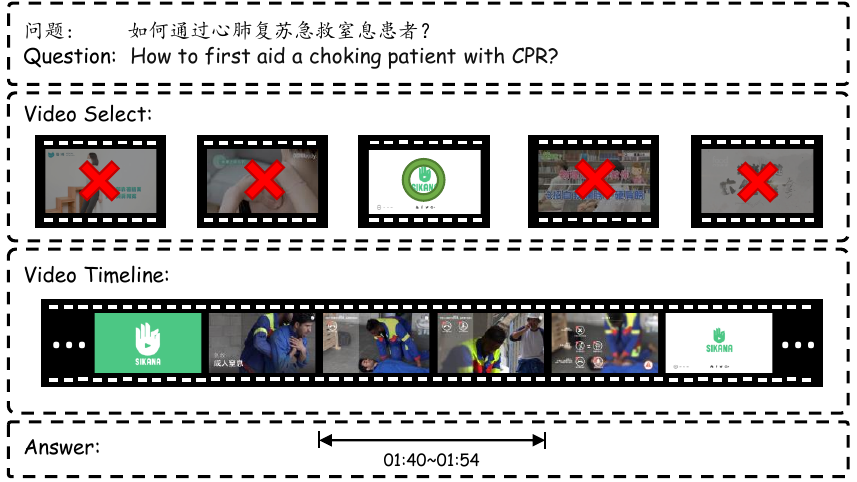}
    \caption{Example of Temporal Answer Grounding with Video Corpus (TAGVC). This figure demonstrates the more challenging TAGVC task. The model must first select the correct video from a corpus of candidates and then localize the relevant temporal segment within that selected video.} \label{Task2}
\end{figure}

Given a medical question and multiple videos, the challenger needs to identify the one that helps answer the question from these videos and clip the sequence of the video corresponds to problem. As shown in Fig. \ref{Task2}.

Assume the number of questions is $n$. The correct answers for question $Q_i$ is video ID $V_i^*$ and video sequence $T_i^*=[t_{start}^*,t_{end}^*] ,i\in[1,n]$.
The challenger provides $50$ answers to the question $Q_i$: $V_{i,j}$, $T_{i,j}=[t_{start},t_{end}]$, $j\in[1,50]$, in the order of confidence from highest to lowest.
The score for question $Q_i$ is:

$$S_{i,j}=\left\{\begin{matrix} 0 & V_{i,j}\neq V_i^* \\ \frac{T_i^* \cap T_{i,j}}{T_i^* \cup T_{i,j}} & V_{i,j}=V_i^* \end{matrix}\right.$$

We use ``$\mathrm{Top-}1/10/50~\mathrm{ IoU}$" and ``Average" as criteria for evaluation. The calculation process is defined as follows:

$$\text{Top-}k \text{ IoU} =\frac{\sum_{i=1}^{n}\max_{j\in[1,k]} S_{i,j}}{n}$$

$$\text{Average}= \frac{\sum_{k=1,10,50}\text{Top-}k \text{ IoU}}{3}$$

            \section{Dataset Statistics and Analysis}
This section presents a detailed statistical analysis of the M$^3$-Med benchmark, quantifying its scale, diversity, and linguistic characteristics.
Tab. \ref{Stat} provides a high-level overview of the dataset's scale, with a breakdown of video and question counts by language (Chinese and English) and reasoning type (simple and complex).

\begin{table}[h]  
                 \centering  
                    \caption{Statistical Overview of the Dataset Size by Language. This table presents the scale of our bilingual dataset, detailing the number of videos, question-answer pairs, and other key metrics for both the Chinese and English subsets.}  
                        \begin{tabular}{cccc}  
    \toprule[2pt]  
\textbf{Language}	&	\textbf{Videos}	&	\textbf{Simple Questions}	&	\textbf{Complex Questions}	\\
\midrule							
\rowcolor{mycolor} \textbf{Chinese}	&	1,628	&	3,208	&	2,746	\\
\textbf{English}	&	2,120	&	3,496	&	3,297	\\
\rowcolor{mycolor} \textbf{Total}	&	3,748	&	6,704	&	6,043	\\

    \bottomrule[2pt]  
\end{tabular}  
                   
                    \label{Stat}  
\end{table}
\begin{figure}[h]
    \centering
    \subfloat[][Topics of Chinese video.]{ 
    \centering
    \includegraphics{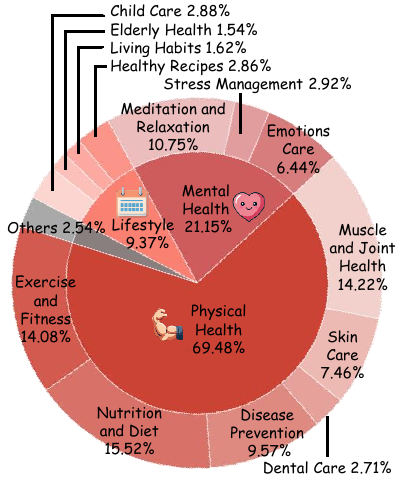}}
    \qquad
    \subfloat[][Topics of English video.]{
    \centering \includegraphics{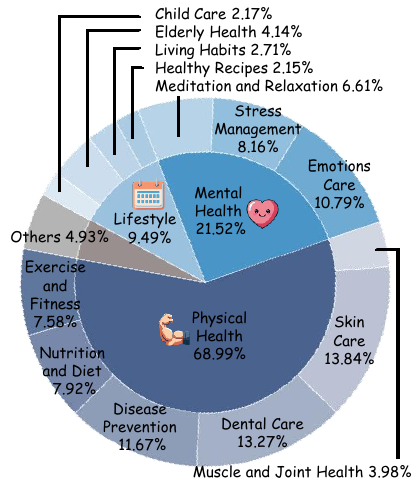}}
    \caption{Distribution of video topics. Those chart visualizes the thematic distribution of the videos in our dataset, demonstrating a broad coverage of medical fields.}
    \label{Themes}
\end{figure}

To showcase the thematic diversity of our dataset, Fig. \ref{Themes} visualizes the topic distribution across both the Chinese and English video subsets.

To further investigate the linguistic distinction between our question types, we analyzed their keyword distributions. After applying the same pre-processing steps used for the subtitles (tokenization for Chinese and lemmatization for English), we compiled the keyword frequencies for simple and complex questions, as presented in Tabs. \ref{ChnCloud} and \ref{EngCloud}.

\begin{table}[h!]
    \caption{
Top keyword frequencies in Chinese simple vs. complex questions.
}
\label{ChnCloud}
\centering
\resizebox{1.0\textwidth}{!}{
\begin{tabular}{lccl}
\toprule[2pt]
		\textbf{Key Word}		&	\textbf{Occ. in S.Q.}	&	\textbf{Occ. in C.Q.}	&		\textbf{Example Sentence}		\\
\midrule												
\rowcolor{mycolor}	\begin{CJK}{UTF8}{gbsn}	如何	\end{CJK}	&	3,127	&	2,915	&	\begin{CJK}{UTF8}{gbsn}	\textcolor{red}{如何}通过脚趾和跟腱的拉伸来纠正扁平足？	\end{CJK}	\\
	\begin{CJK}{UTF8}{gbsn}	通过	\end{CJK}	&	586	&	585	&	\begin{CJK}{UTF8}{gbsn}	如何\textcolor{red}{通过}正确有效地拍打肩膀的次数来疏通结节肩？	\end{CJK}	\\
\rowcolor{mycolor}	\begin{CJK}{UTF8}{gbsn}	进行	\end{CJK}	&	469	&	478	&	\begin{CJK}{UTF8}{gbsn}	如何\textcolor{red}{进行}髋关节旋转来使髋关节灵活？	\end{CJK}	\\
	\begin{CJK}{UTF8}{gbsn}	缓解	\end{CJK}	&	374	&	353	&	\begin{CJK}{UTF8}{gbsn}	如何站立把腿往外拉伸\textcolor{red}{缓解}腰疼？	\end{CJK}	\\
\rowcolor{mycolor}	\begin{CJK}{UTF8}{gbsn}	运动	\end{CJK}	&	311	&	337	&	\begin{CJK}{UTF8}{gbsn}	如何通过站立侧抬\textcolor{red}{运动}来保养髋关节？	\end{CJK}	\\
	\begin{CJK}{UTF8}{gbsn}	放松	\end{CJK}	&	279	&	255	&	\begin{CJK}{UTF8}{gbsn}	如何拉伸三头肌来\textcolor{red}{放松}肩膀？	\end{CJK}	\\
\rowcolor{mycolor}	\begin{CJK}{UTF8}{gbsn}	训练	\end{CJK}	&	206	&	189	&	\begin{CJK}{UTF8}{gbsn}	如何卧倒通过简单抬起大腿\textcolor{red}{训练}髂腰肌？	\end{CJK}	\\
	\begin{CJK}{UTF8}{gbsn}	疼痛	\end{CJK}	&	184	&	174	&	\begin{CJK}{UTF8}{gbsn}	一个人如何使用泡沫滚轴改善肩胛骨和后背\textcolor{red}{疼痛}？	\end{CJK}	\\
\rowcolor{mycolor}	\begin{CJK}{UTF8}{gbsn}	拉伸	\end{CJK}	&	173	&	290	&	\begin{CJK}{UTF8}{gbsn}	如何用手支撑身体然后头和腿向两边打开\textcolor{red}{拉伸}身体？	\end{CJK}	\\
	\begin{CJK}{UTF8}{gbsn}	肌肉	\end{CJK}	&	166	&	253	&	\begin{CJK}{UTF8}{gbsn}	如何使用大腿后侧跟屁股的\textcolor{red}{肌肉}矫正走路姿势？	\end{CJK}	\\
\rowcolor{mycolor}	\begin{CJK}{UTF8}{gbsn}	使用	\end{CJK}	&	158	&	95	&	\begin{CJK}{UTF8}{gbsn}	一个人如何\textcolor{red}{使用}泡沫滚轴改善肩胛骨和后背疼痛？	\end{CJK}	\\
	\begin{CJK}{UTF8}{gbsn}	伸展	\end{CJK}	&	150	&	175	&	\begin{CJK}{UTF8}{gbsn}	如何进行胸肌静态躯干\textcolor{red}{伸展}？	\end{CJK}	\\
\rowcolor{mycolor}	\begin{CJK}{UTF8}{gbsn}	锻炼	\end{CJK}	&	136	&	191	&	\begin{CJK}{UTF8}{gbsn}	如何通过爬行训练\textcolor{red}{锻炼}核心？	\end{CJK}	\\
	\begin{CJK}{UTF8}{gbsn}	利用	\end{CJK}	&	121	&	98	&	\begin{CJK}{UTF8}{gbsn}	如何\textcolor{red}{利用}桌子强化肩颈肌力与颈椎活动力？	\end{CJK}	\\
\rowcolor{mycolor}	\begin{CJK}{UTF8}{gbsn}	按摩	\end{CJK}	&	112	&	167	&	\begin{CJK}{UTF8}{gbsn}	如何一拍二抓三\textcolor{red}{按摩}来使膝盖不痛？	\end{CJK}	\\
	\begin{CJK}{UTF8}{gbsn}	膝盖	\end{CJK}	&	103	&	112	&	\begin{CJK}{UTF8}{gbsn}	如何下蹲\textcolor{red}{膝盖}弯曲缓解腰疼？	\end{CJK}	\\
\rowcolor{mycolor}	\begin{CJK}{UTF8}{gbsn}	动作	\end{CJK}	&	102	&	177	&	\begin{CJK}{UTF8}{gbsn}	如何将一只脚踩在另一只脚上，进行外旋\textcolor{red}{动作}来纠正扁平足？	\end{CJK}	\\
	\begin{CJK}{UTF8}{gbsn}	正确	\end{CJK}	&	101	&	317	&	\begin{CJK}{UTF8}{gbsn}	如何\textcolor{red}{正确}使用工具正反腕来训练手腕？	\end{CJK}	\\
\rowcolor{mycolor}	\begin{CJK}{UTF8}{gbsn}	大腿	\end{CJK}	&	97	&	121	&	\begin{CJK}{UTF8}{gbsn}	如何卧倒通过简单抬起\textcolor{red}{大腿}训练髂腰肌？	\end{CJK}	\\
	\begin{CJK}{UTF8}{gbsn}	肩膀	\end{CJK}	&	87	&	176	&	\begin{CJK}{UTF8}{gbsn}	手臂无力体虚，怎么拍打\textcolor{red}{肩膀}两侧，使结节肩疏通？	\end{CJK}	\\
\bottomrule[2pt]
\end{tabular}
}
\begin{tablenotes}
        \footnotesize
        \item[*] Occ. in S.Q. (Occurrences times in simple questions); Occ. in C.Q. (Occurrences times in complex questions).
\end{tablenotes}
\end{table}

\begin{table}[h!]
    \caption{
Top keyword frequencies in English simple vs. complex questions.
}
\label{EngCloud}
\centering
\resizebox{1.0\textwidth}{!}{
\begin{tabular}{lccl}
\toprule[2pt]
	\textbf{Key Word}		&	\textbf{Occ. in S.Q.}	&	\textbf{Occ. in C.Q.}	&		\textbf{Example Sentence}		\\
\midrule								
\rowcolor{mycolor}	how	&	3,469	&	3,260	&	\textcolor{red}{How} to install a watertight seal using a philips screwdriver?
	\\
	use	&	435	&	737	&	How to \textcolor{red}{use} a water flosser?	\\
\rowcolor{mycolor}	make	&	299	&	522	&	How to \textcolor{red}{make} protein pancakes without using protein powder?	\\
	treat	&	235	&	369	&	How to use a bite guard to \textcolor{red}{treat} clenching jaws?	\\
\rowcolor{mycolor}	get	&	176	&	536	&	How to \textcolor{red}{get} rid of pain and joint dysfunction at home?	\\
	using	&	168	&	387	&	How to create a bluesy sound \textcolor{red}{using} a minor blue scale?	\\
\rowcolor{mycolor}	clean	&	143	&	167	&	How to \textcolor{red}{clean} the area around the implant?	\\
	remove	&	143	&	263	&	How to \textcolor{red}{remove} TMJ from the jaw?	\\
\rowcolor{mycolor}	perform	&	119	&	246	&	How to \textcolor{red}{perform} a double chin exercise while lying down to strengthen the shoulder muscles?	\\
	apply	&	119	&	178	&	How to \textcolor{red}{apply} ice on the face after rhinoplasty?	\\
\rowcolor{mycolor}	rid	&	93	&	311	&	How to get \textcolor{red}{rid} of pain and joint dysfunction at home?	\\
	create	&	88	&	128	&	How to \textcolor{red}{create} a bluesy sound using a minor blue scale?	\\
\rowcolor{mycolor}	identify	&	83	&	335	&	How to \textcolor{red}{identify} a urologist?	\\
	pain	&	73	&	80	&	How to perform a double chin exercise to fix and neck \textcolor{red}{pain}?	\\
\rowcolor{mycolor}	tooth	&	71	&	89	&	How to treat impacted canine \textcolor{red}{teeth}?	\\
	heart	&	69	&	117	&	How to measure a baby's \textcolor{red}{heart} girth?	\\
\rowcolor{mycolor}	prepare	&	65	&	194	&	How to \textcolor{red}{prepare} a skillet for egg white omelets?	\\
	prevent	&	63	&	112	&	How to use a mercury compression device to \textcolor{red}{prevent} venous thrombosis?	\\
\rowcolor{mycolor}	check	&	61	&	178	&	How to \textcolor{red}{check} if a fuse is blown?	\\
	hair	&	59	&	79	&	How to remove chewing gum from \textcolor{red}{hair}?	\\
\bottomrule[2pt]
\end{tabular}
}
\begin{tablenotes}
        \footnotesize
        \item[*] Occ. in S.Q. (Occurrences times in simple questions); Occ. in C.Q. (Occurrences times in complex questions).
\end{tablenotes}
\end{table}

            \section{Experiments}
To validate the challenges posed by our benchmark and establish strong baseline performances, we conducted a comprehensive set of experiments evaluating both state-of-the-art models and human challengers.

\begin{table}[h]  
    \def\valAAAAAA{0.5737}	\def\valAAAABA{0.4094}	\def\valAAAACA{0.2206}	\def\valAAAADA{0.4012}	\def\valAAABAA{0.5284}	\def\valAAABBA{\textbf{0.3738}}	\def\valAAABCA{0.2368}	\def\valAAABDA{0.3782}	
\def\valAAAAAB{0.6237}	\def\valAAAABB{0.4493}	\def\valAAAACB{0.2764}	\def\valAAAADB{\textbf{0.4498}}	\def\valAAABAB{0.5205}	\def\valAAABBB{0.3679}	\def\valAAABCB{0.2368}	\def\valAAABDB{0.3788}	
\def\valAAAAAD{\textbf{0.6604}}	\def\valAAAABD{\textbf{0.5034}}	\def\valAAAACD{\textbf{0.2829}}	\def\valAAAADD{0.4363}	\def\valAAABAD{\textbf{0.5503}}	\def\valAAABBD{0.3727}	\def\valAAABCD{\textbf{0.2881}}	\def\valAAABDD{\textbf{0.4123}}	
\def\valAABAAA{0.5557}	\def\valAABABA{0.3894}	\def\valAABACA{0.2239}	\def\valAABADA{0.3891}	\def\valAABBAA{0.4990}	\def\valAABBBA{0.3738}	\def\valAABBCA{0.2133}	\def\valAABBDA{0.3592}	
\def\valAABAAB{0.6164}	\def\valAABABB{0.4325}	\def\valAABACB{0.2681}	\def\valAABADB{0.4390}	\def\valAABBAB{0.5068}	\def\valAABBBB{0.3581}	\def\valAABBCB{0.2446}	\def\valAABBDB{0.3768}	
\def\valAABAAD{\textbf{0.6232}}	\def\valAABABD{\textbf{0.4955}}	\def\valAABACD{\textbf{0.2878}}	\def\valAABADD{\textbf{0.4470}}	\def\valAABBAD{\textbf{0.5311}}	\def\valAABBBD{\textbf{0.4595}}	\def\valAABBCD{\textbf{0.2551}}	\def\valAABBDD{\textbf{0.4540}}	
\def\valAACAAA{0.2333}	\def\valAACABA{0.1475}	\def\valAACACA{0.0583}	\def\valAACADA{0.1519}	\def\valAACBAA{0.2298}	\def\valAACBBA{0.1453}	\def\valAACBCA{0.0575}	\def\valAACBDA{0.1497}	
\def\valAACAAB{0.2365}	\def\valAACABB{0.1411}	\def\valAACACB{0.0586}	\def\valAACADB{0.1556}	\def\valAACBAB{0.2330}	\def\valAACBBB{0.1391}	\def\valAACBCB{0.0578}	\def\valAACBDB{0.1533}	
\def\valAACAAC{\textbf{0.2799}}	\def\valAACABC{\textbf{0.1492}}	\def\valAACACC{\textbf{0.0747}}	\def\valAACADC{\textbf{0.1699}}	\def\valAACBAC{\textbf{0.2747}}	\def\valAACBBC{\textbf{0.1787}}	\def\valAACBCC{\textbf{0.1031}}	\def\valAACBDC{\textbf{0.1764}}	
\def\valAACAAD{0.2420}	\def\valAACABD{0.1452}	\def\valAACACD{0.0581}	\def\valAACADD{0.1517}	\def\valAACBAD{0.2383}	\def\valAACBBD{0.1430}	\def\valAACBCD{0.0572}	\def\valAACBDD{0.1494}	
\def\valAADAAA{0.0770}	\def\valAADABA{\textbf{0.0459}}	\def\valAADACA{0.0141}	\def\valAADADA{0.0379}	\def\valAADBAA{0.0585}	\def\valAADBBA{0.0375}	\def\valAADBCA{0.0160}	\def\valAADBDA{\textbf{0.0488}}	
\def\valAADAAB{\textbf{0.0789}}	\def\valAADABB{0.0436}	\def\valAADACB{0.0152}	\def\valAADADB{0.0413}	\def\valAADBAB{\textbf{0.0679}}	\def\valAADBBB{0.0433}	\def\valAADBCB{0.0166}	\def\valAADBDB{0.0454}	
\def\valAADAAC{0.0627}	\def\valAADABC{0.0405}	\def\valAADACC{0.0183}	\def\valAADADC{\textbf{0.0460}}	\def\valAADBAC{0.0575}	\def\valAADBBC{0.0429}	\def\valAADBCC{0.0189}	\def\valAADBDC{0.0461}	
\def\valAADAAD{0.0665}	\def\valAADABD{0.0453}	\def\valAADACD{\textbf{0.0185}}	\def\valAADADD{0.0382}	\def\valAADBAD{0.0657}	\def\valAADBBD{\textbf{0.0491}}	\def\valAADBCD{\textbf{0.0198}}	\def\valAADBDD{0.0427}	
\def\valABAAAA{0.5671}	\def\valABAABA{0.4065}	\def\valABAACA{0.2358}	\def\valABAADA{0.4028}	\def\valABABAA{\textbf{0.5168}}	\def\valABABBA{0.3656}	\def\valABABCA{0.2316}	\def\valABABDA{0.3699}	
\def\valABAAAB{\textbf{0.6052}}	\def\valABAABB{0.4313}	\def\valABAACB{0.2658}	\def\valABAADB{0.4341}	\def\valABABAB{0.5090}	\def\valABABBB{0.3598}	\def\valABABCB{0.2316}	\def\valABABDB{0.3705}	
\def\valABAAAD{0.5951}	\def\valABAABD{\textbf{0.5085}}	\def\valABAACD{\textbf{0.3094}}	\def\valABAADD{\textbf{0.4752}}	\def\valABABAD{0.5132}	\def\valABABBD{\textbf{0.4418}}	\def\valABABCD{\textbf{0.2943}}	\def\valABABDD{\textbf{0.3997}}	
\def\valABBAAA{0.5284}	\def\valABBABA{0.3816}	\def\valABBACA{0.2290}	\def\valABBADA{0.3866}	\def\valABBBAA{0.4880}	\def\valABBBBA{0.3656}	\def\valABBBCA{0.2086}	\def\valABBBDA{0.3513}	
\def\valABBAAB{0.5818}	\def\valABBABB{\textbf{0.4217}}	\def\valABBACB{\textbf{0.2571}}	\def\valABBADB{0.4202}	\def\valABBBAB{0.4957}	\def\valABBBBB{0.3502}	\def\valABBBCB{0.2392}	\def\valABBBDB{0.3685}	
\def\valABBAAD{\textbf{0.6499}}	\def\valABBABD{0.4210}	\def\valABBACD{0.2459}	\def\valABBADD{\textbf{0.4881}}	\def\valABBBAD{\textbf{0.5625}}	\def\valABBBBD{\textbf{0.4363}}	\def\valABBBCD{\textbf{0.2981}}	\def\valABBBDD{\textbf{0.4070}}	
\def\valABCAAA{0.2323}	\def\valABCABA{0.1452}	\def\valABCACA{0.0546}	\def\valABCADA{0.1496}	\def\valABCBAA{0.2288}	\def\valABCBBA{0.1430}	\def\valABCBCA{0.0538}	\def\valABCBDA{0.1474}	
\def\valABCAAB{0.2355}	\def\valABCABB{0.1387}	\def\valABCACB{0.0549}	\def\valABCADB{0.1534}	\def\valABCBAB{0.2320}	\def\valABCBBB{0.1366}	\def\valABCBCB{0.0541}	\def\valABCBDB{0.1511}	
\def\valABCAAC{\textbf{0.2485}}	\def\valABCABC{\textbf{0.1901}}	\def\valABCACC{\textbf{0.0934}}	\def\valABCADC{\textbf{0.2018}}	\def\valABCBAC{\textbf{0.2625}}	\def\valABCBBC{\textbf{0.1693}}	\def\valABCBCC{\textbf{0.0978}}	\def\valABCBDC{\textbf{0.1579}}	
\def\valABCAAD{0.2428}	\def\valABCABD{0.1475}	\def\valABCACD{0.0617}	\def\valABCADD{0.1539}	\def\valABCBAD{0.2393}	\def\valABCBBD{0.1453}	\def\valABCBCD{0.0609}	\def\valABCBDD{0.1517}	
\def\valABDAAA{\textbf{0.0783}}	\def\valABDABA{0.0444}	\def\valABDACA{0.0172}	\def\valABDADA{0.0405}	\def\valABDBAA{0.0671}	\def\valABDBBA{0.0445}	\def\valABDBCA{\textbf{0.0181}}	\def\valABDBDA{\textbf{0.0410}}	
\def\valABDAAB{0.0696}	\def\valABDABB{0.0455}	\def\valABDACB{\textbf{0.0181}}	\def\valABDADB{0.0438}	\def\valABDBAB{0.0688}	\def\valABDBBB{\textbf{0.0464}}	\def\valABDBCB{0.0166}	\def\valABDBDB{0.0393}	
\def\valABDAAC{0.0712}	\def\valABDABC{0.0427}	\def\valABDACC{0.0154}	\def\valABDADC{\textbf{0.0484}}	\def\valABDBAC{0.0693}	\def\valABDBBC{0.0412}	\def\valABDBCC{0.0158}	\def\valABDBDC{0.0410}	
\def\valABDAAD{0.0741}	\def\valABDABD{\textbf{0.0470}}	\def\valABDACD{0.0151}	\def\valABDADD{0.0389}	\def\valABDBAD{\textbf{0.0700}}	\def\valABDBBD{0.0420}	\def\valABDBCD{0.0180}	\def\valABDBDD{0.0403}	

\centering
\caption{
The results of task 1 with different settings.
}  
\resizebox{1.0\textwidth}{!}{
\begin{tabular}{llcccccccc}  
\toprule[2pt]
\multirow{2}{*}{\hspace{1mm}\textbf{Type}}	&		\multirow{2}{*}{\hspace{2mm}\textbf{Model}}	&		\multicolumn{4}{c}{\textbf{Simple Questions}}										&		\multicolumn{4}{c}{\textbf{Complex Questions}}										\\
						\cmidrule(lr){3-6}												\cmidrule(lr){7-10}										
	&			&		$\mathbf{\tau\textbf{=}0.3}$	&		$\mathbf{\tau\textbf{=}0.5}$	&		$\mathbf{\tau\textbf{=}0.7}$	&		\textbf{mIoU}	&		$\mathbf{\tau\textbf{=}0.3}$	&		$\mathbf{\tau\textbf{=}0.5}$	&		$\mathbf{\tau\textbf{=}0.7}$	&		\textbf{mIoU}	\\
\midrule
\multicolumn{10}{c}{\textbf{Chinese Questions}}	\\
\midrule
\cellcolor{mycolor} \textbf{Vid. + Sub. + K.G.}	&	\cellcolor{mycolor}	$\begin{array}{l} \text{MutualSL \cite{MutualSL}} \\ \text{PBL \cite{PBL}} \\ \text{Qwen-2.5-VL \cite{Qwen}} \end{array}$	&	\cellcolor{mycolor}	$\begin{array}{c} \valAAAAAA \\ \valAAAAAB \\ \valAAAAAD \end{array}$	&	\cellcolor{mycolor}	$\begin{array}{c} \valAAAABA \\ \valAAAABB \\ \valAAAABD \end{array}$	&	\cellcolor{mycolor}	$\begin{array}{c} \valAAAACA \\ \valAAAACB \\ \valAAAACD \end{array}$	&	\cellcolor{mycolor}	$\begin{array}{c} \valAAAADA \\ \valAAAADB \\ \valAAAADD \end{array}$	&	\cellcolor{mycolor}	$\begin{array}{c} \valAAABAA \\ \valAAABAB \\ \valAAABAD \end{array}$	&	\cellcolor{mycolor}	$\begin{array}{c} \valAAABBA \\ \valAAABBB \\ \valAAABBD \end{array}$	&	\cellcolor{mycolor}	$\begin{array}{c} \valAAABCA \\ \valAAABCB \\ \valAAABCD \end{array}$	&	\cellcolor{mycolor}	$\begin{array}{c} \valAAABDA \\ \valAAABDB \\ \valAAABDD \end{array}$	\\
\cellcolor{baise} \textbf{Vid. + Sub.}	&	\cellcolor{baise}	$\begin{array}{l} \text{MutualSL \cite{MutualSL}} \\ \text{PBL \cite{PBL}} \\ \text{Qwen-2.5-VL \cite{Qwen}} \end{array}$	&	\cellcolor{baise}	$\begin{array}{c} \valAABAAA \\ \valAABAAB \\ \valAABAAD \end{array}$	&	\cellcolor{baise}	$\begin{array}{c} \valAABABA \\ \valAABABB \\ \valAABABD \end{array}$	&	\cellcolor{baise}	$\begin{array}{c} \valAABACA \\ \valAABACB \\ \valAABACD \end{array}$	&	\cellcolor{baise}	$\begin{array}{c} \valAABADA \\ \valAABADB \\ \valAABADD \end{array}$	&	\cellcolor{baise}	$\begin{array}{c} \valAABBAA \\ \valAABBAB \\ \valAABBAD \end{array}$	&	\cellcolor{baise}	$\begin{array}{c} \valAABBBA \\ \valAABBBB \\ \valAABBBD \end{array}$	&	\cellcolor{baise}	$\begin{array}{c} \valAABBCA \\ \valAABBCB \\ \valAABBCD \end{array}$	&	\cellcolor{baise}	$\begin{array}{c} \valAABBDA \\ \valAABBDB \\ \valAABBDD \end{array}$	\\
\cellcolor{mycolor} \textbf{Sub. + K.G.}	&	\cellcolor{mycolor}	$\begin{array}{l} \text{MutualSL \cite{MutualSL}} \\ \text{PBL \cite{PBL}} \\ \text{GPT-4o \cite{GPT}} \\ \text{Qwen-2.5-VL \cite{Qwen}} \end{array}$	&	\cellcolor{mycolor}	$\begin{array}{c} \valAACAAA \\ \valAACAAB \\ \valAACAAC \\ \valAACAAD \end{array}$	&	\cellcolor{mycolor}	$\begin{array}{c} \valAACABA \\ \valAACABB \\ \valAACABC \\ \valAACABD \end{array}$	&	\cellcolor{mycolor}	$\begin{array}{c} \valAACACA \\ \valAACACB \\ \valAACACC \\ \valAACACD \end{array}$	&	\cellcolor{mycolor}	$\begin{array}{c} \valAACADA \\ \valAACADB \\ \valAACADC \\ \valAACADD \end{array}$	&	\cellcolor{mycolor}	$\begin{array}{c} \valAACBAA \\ \valAACBAB \\ \valAACBAC \\ \valAACBAD \end{array}$	&	\cellcolor{mycolor}	$\begin{array}{c} \valAACBBA \\ \valAACBBB \\ \valAACBBC \\ \valAACBBD \end{array}$	&	\cellcolor{mycolor}	$\begin{array}{c} \valAACBCA \\ \valAACBCB \\ \valAACBCC \\ \valAACBCD \end{array}$	&	\cellcolor{mycolor}	$\begin{array}{c} \valAACBDA \\ \valAACBDB \\ \valAACBDC \\ \valAACBDD \end{array}$	\\
\cellcolor{baise} \textbf{Sub.}	&	\cellcolor{baise}	$\begin{array}{l} \text{MutualSL \cite{MutualSL}} \\ \text{PBL \cite{PBL}} \\ \text{GPT-4o \cite{GPT}} \\ \text{Qwen-2.5-VL \cite{Qwen}} \end{array}$	&	\cellcolor{baise}	$\begin{array}{c} \valAADAAA \\ \valAADAAB \\ \valAADAAC \\ \valAADAAD \end{array}$	&	\cellcolor{baise}	$\begin{array}{c} \valAADABA \\ \valAADABB \\ \valAADABC \\ \valAADABD \end{array}$	&	\cellcolor{baise}	$\begin{array}{c} \valAADACA \\ \valAADACB \\ \valAADACC \\ \valAADACD \end{array}$	&	\cellcolor{baise}	$\begin{array}{c} \valAADADA \\ \valAADADB \\ \valAADADC \\ \valAADADD \end{array}$	&	\cellcolor{baise}	$\begin{array}{c} \valAADBAA \\ \valAADBAB \\ \valAADBAC \\ \valAADBAD \end{array}$	&	\cellcolor{baise}	$\begin{array}{c} \valAADBBA \\ \valAADBBB \\ \valAADBBC \\ \valAADBBD \end{array}$	&	\cellcolor{baise}	$\begin{array}{c} \valAADBCA \\ \valAADBCB \\ \valAADBCC \\ \valAADBCD \end{array}$	&	\cellcolor{baise}	$\begin{array}{c} \valAADBDA \\ \valAADBDB \\ \valAADBDC \\ \valAADBDD \end{array}$	\\
\cellcolor{mycolor} \textbf{Human} & \cellcolor{mycolor} \hspace{1cm} \textbf{---}  & \cellcolor{mycolor} $\begin{array}{c} 0.8657 \end{array}$ & \cellcolor{mycolor} $\begin{array}{c} 0.7334 \end{array}$ & \cellcolor{mycolor} $\begin{array}{c} 0.7129 \end{array}$ & \cellcolor{mycolor} $\begin{array}{c} 0.7940 \end{array}$ & \cellcolor{mycolor} $\begin{array}{c} 0.8720 \end{array}$ & \cellcolor{mycolor} $\begin{array}{c} 0.7260 \end{array}$ & \cellcolor{mycolor} $\begin{array}{c} 0.7173 \end{array}$ & \cellcolor{mycolor} $\begin{array}{c} 0.8013 \end{array}$ \\
\midrule
\multicolumn{10}{c}{\textbf{English Questions}}	\\
\midrule
\cellcolor{mycolor} \textbf{Vid. + Sub. + K.G.}	&	\cellcolor{mycolor}	$\begin{array}{l} \text{MutualSL \cite{MutualSL}} \\ \text{PBL \cite{PBL}} \\ \text{Qwen-2.5-VL \cite{Qwen}} \end{array}$	&	\cellcolor{mycolor}	$\begin{array}{c} \valABAAAA \\ \valABAAAB \\ \valABAAAD \end{array}$	&	\cellcolor{mycolor}	$\begin{array}{c} \valABAABA \\ \valABAABB \\ \valABAABD \end{array}$	&	\cellcolor{mycolor}	$\begin{array}{c} \valABAACA \\ \valABAACB \\ \valABAACD \end{array}$	&	\cellcolor{mycolor}	$\begin{array}{c} \valABAADA \\ \valABAADB \\ \valABAADD \end{array}$	&	\cellcolor{mycolor}	$\begin{array}{c} \valABABAA \\ \valABABAB \\ \valABABAD \end{array}$	&	\cellcolor{mycolor}	$\begin{array}{c} \valABABBA \\ \valABABBB \\ \valABABBD \end{array}$	&	\cellcolor{mycolor}	$\begin{array}{c} \valABABCA \\ \valABABCB \\ \valABABCD \end{array}$	&	\cellcolor{mycolor}	$\begin{array}{c} \valABABDA \\ \valABABDB \\ \valABABDD \end{array}$	\\
\cellcolor{baise} \textbf{Vid. + Sub.}	&	\cellcolor{baise}	$\begin{array}{l} \text{MutualSL \cite{MutualSL}} \\ \text{PBL \cite{PBL}} \\ \text{Qwen-2.5-VL \cite{Qwen}} \end{array}$	&	\cellcolor{baise}	$\begin{array}{c} \valABBAAA \\ \valABBAAB \\ \valABBAAD \end{array}$	&	\cellcolor{baise}	$\begin{array}{c} \valABBABA \\ \valABBABB \\ \valABBABD \end{array}$	&	\cellcolor{baise}	$\begin{array}{c} \valABBACA \\ \valABBACB \\ \valABBACD \end{array}$	&	\cellcolor{baise}	$\begin{array}{c} \valABBADA \\ \valABBADB \\ \valABBADD \end{array}$	&	\cellcolor{baise}	$\begin{array}{c} \valABBBAA \\ \valABBBAB \\ \valABBBAD \end{array}$	&	\cellcolor{baise}	$\begin{array}{c} \valABBBBA \\ \valABBBBB \\ \valABBBBD \end{array}$	&	\cellcolor{baise}	$\begin{array}{c} \valABBBCA \\ \valABBBCB \\ \valABBBCD \end{array}$	&	\cellcolor{baise}	$\begin{array}{c} \valABBBDA \\ \valABBBDB \\ \valABBBDD \end{array}$	\\
\cellcolor{mycolor} \textbf{Sub. + K.G.}	&	\cellcolor{mycolor}	$\begin{array}{l} \text{MutualSL \cite{MutualSL}} \\ \text{PBL \cite{PBL}} \\ \text{GPT-4o \cite{GPT}} \\ \text{Qwen-2.5-VL \cite{Qwen}} \end{array}$	&	\cellcolor{mycolor}	$\begin{array}{c} \valABCAAA \\ \valABCAAB \\ \valABCAAC \\ \valABCAAD \end{array}$	&	\cellcolor{mycolor}	$\begin{array}{c} \valABCABA \\ \valABCABB \\ \valABCABC \\ \valABCABD \end{array}$	&	\cellcolor{mycolor}	$\begin{array}{c} \valABCACA \\ \valABCACB \\ \valABCACC \\ \valABCACD \end{array}$	&	\cellcolor{mycolor}	$\begin{array}{c} \valABCADA \\ \valABCADB \\ \valABCADC \\ \valABCADD \end{array}$	&	\cellcolor{mycolor}	$\begin{array}{c} \valABCBAA \\ \valABCBAB \\ \valABCBAC \\ \valABCBAD \end{array}$	&	\cellcolor{mycolor}	$\begin{array}{c} \valABCBBA \\ \valABCBBB \\ \valABCBBC \\ \valABCBBD \end{array}$	&	\cellcolor{mycolor}	$\begin{array}{c} \valABCBCA \\ \valABCBCB \\ \valABCBCC \\ \valABCBCD \end{array}$	&	\cellcolor{mycolor}	$\begin{array}{c} \valABCBDA \\ \valABCBDB \\ \valABCBDC \\ \valABCBDD \end{array}$	\\
\cellcolor{baise} \textbf{Sub.}	&	\cellcolor{baise}	$\begin{array}{l} \text{MutualSL \cite{MutualSL}} \\ \text{PBL \cite{PBL}} \\ \text{GPT-4o \cite{GPT}} \\ \text{Qwen-2.5-VL \cite{Qwen}} \end{array}$	&	\cellcolor{baise}	$\begin{array}{c} \valABDAAA \\ \valABDAAB \\ \valABDAAC \\ \valABDAAD \end{array}$	&	\cellcolor{baise}	$\begin{array}{c} \valABDABA \\ \valABDABB \\ \valABDABC \\ \valABDABD \end{array}$	&	\cellcolor{baise}	$\begin{array}{c} \valABDACA \\ \valABDACB \\ \valABDACC \\ \valABDACD \end{array}$	&	\cellcolor{baise}	$\begin{array}{c} \valABDADA \\ \valABDADB \\ \valABDADC \\ \valABDADD \end{array}$	&	\cellcolor{baise}	$\begin{array}{c} \valABDBAA \\ \valABDBAB \\ \valABDBAC \\ \valABDBAD \end{array}$	&	\cellcolor{baise}	$\begin{array}{c} \valABDBBA \\ \valABDBBB \\ \valABDBBC \\ \valABDBBD \end{array}$	&	\cellcolor{baise}	$\begin{array}{c} \valABDBCA \\ \valABDBCB \\ \valABDBCC \\ \valABDBCD \end{array}$	&	\cellcolor{baise}	$\begin{array}{c} \valABDBDA \\ \valABDBDB \\ \valABDBDC \\ \valABDBDD \end{array}$	\\
\cellcolor{mycolor} \textbf{Human} & \cellcolor{mycolor} \hspace{1cm} \textbf{---}  & \cellcolor{mycolor} $\begin{array}{c} 0.8240 \end{array}$ & \cellcolor{mycolor} $\begin{array}{c} 0.6907 \end{array}$ & \cellcolor{mycolor} $\begin{array}{c} 0.6860 \end{array}$ & \cellcolor{mycolor} $\begin{array}{c} 0.7330 \end{array}$ & \cellcolor{mycolor} $\begin{array}{c} 0.8181 \end{array}$ & \cellcolor{mycolor} $\begin{array}{c} 0.7158 \end{array}$ & \cellcolor{mycolor} $\begin{array}{c} 0.6873 \end{array}$ & \cellcolor{mycolor} $\begin{array}{c} 0.7287 \end{array}$ \\
            \bottomrule[2pt]  
        \end{tabular}    
    }
    \label{result1}  
\begin{tablenotes}
\footnotesize
\item[*] \textbf{Bold} represents the grade of the top-performing model in the current setting.
$\tau=0.3$: (Score of $\mathrm{IoU}_{\tau=0.3}$);
$\tau=0.5$: (Score of $\mathrm{IoU}_{\tau=0.5}$);
$\tau=0.7$: (Score of $\mathrm{IoU}_{\tau=0.3}$);
mIoU: (The mean IoU of all sample);
Type: (The type of settings when challengers solving the problem);
Vid.: (Videos);
Sub.: (Subtitles in text format);
K.G.: (Knowledge graphs in text format);
Human: (Tasks solved manually).
\end{tablenotes}

\end{table}

\begin{table}[h]  
    \def\valBAAAAA{0.2240}\def\valBAAABA{0.3810}\def\valBAAACA{0.4779}\def\valBAAADA{0.3610}\def\valBAABAA{0.1378}\def\valBAABBA{0.2316}\def\valBAABCA{0.3032}\def\valBAABDA{0.2242}
\def\valBAAAAB{0.2364}\def\valBAAABB{\textbf{0.4376}}\def\valBAAACB{\textbf{0.5633}}\def\valBAAADB{0.4124}\def\valBAABAB{0.1454}\def\valBAABBB{0.2659}\def\valBAABCB{\textbf{0.3572}}\def\valBAABDB{0.2562}
\def\valBAAAAD{\textbf{0.2761}}\def\valBAAABD{0.4339}\def\valBAAACD{0.5408}\def\valBAAADD{\textbf{0.4169}}\def\valBAABAD{\textbf{0.2372}}\def\valBAABBD{\textbf{0.2865}}\def\valBAABCD{0.3408}\def\valBAABDD{\textbf{0.2882}}
\def\valBABAAA{0.2221}\def\valBABABA{0.3799}\def\valBABACA{0.4773}\def\valBABADA{0.3598}\def\valBABBAA{0.1355}\def\valBABBBA{0.2298}\def\valBABBCA{0.3018}\def\valBABBDA{0.2224}
\def\valBABAAB{0.2346}\def\valBABABB{\textbf{0.4368}}\def\valBABACB{\textbf{0.5631}}\def\valBABADB{0.4115}\def\valBABBAB{0.1431}\def\valBABBBB{\textbf{0.2642}}\def\valBABBCB{\textbf{0.3560}}\def\valBABBDB{0.2544}
\def\valBABAAD{\textbf{0.2619}}\def\valBABABD{0.4211}\def\valBABACD{0.5592}\def\valBABADD{\textbf{0.4141}}\def\valBABBAD{\textbf{0.2202}}\def\valBABBBD{0.2615}\def\valBABBCD{0.3437}\def\valBABBDD{\textbf{0.2751}}
\def\valBACAAA{0.0894}\def\valBACABA{0.1618}\def\valBACACA{0.2148}\def\valBACADA{0.1553}\def\valBACBAA{0.0545}\def\valBACBBA{0.0979}\def\valBACBCA{0.1358}\def\valBACBDA{0.0961}
\def\valBACAAB{0.0974}\def\valBACABB{0.1699}\def\valBACACB{0.2205}\def\valBACADB{0.1626}\def\valBACBAB{0.0594}\def\valBACBBB{0.1028}\def\valBACBCB{0.1394}\def\valBACBDB{0.1005}
\def\valBACAAC{\textbf{0.1021}}\def\valBACABC{\textbf{0.2067}}\def\valBACACC{\textbf{0.2575}}\def\valBACADC{\textbf{0.1888}}\def\valBACBAC{\textbf{0.0771}}\def\valBACBBC{\textbf{0.1178}}\def\valBACBCC{\textbf{0.1772}}\def\valBACBDC{\textbf{0.1240}}
\def\valBACAAD{0.1000}\def\valBACABD{0.1718}\def\valBACACD{0.2249}\def\valBACADD{0.1656}\def\valBACBAD{0.0610}\def\valBACBBD{0.1039}\def\valBACBCD{0.1422}\def\valBACBDD{0.1024}
\def\valBADAAA{\textbf{0.0797}}\def\valBADABA{\textbf{0.0492}}\def\valBADACA{0.0171}\def\valBADADA{\textbf{0.0487}}\def\valBADBAA{\textbf{0.0715}}\def\valBADBBA{\textbf{0.0454}}\def\valBADBCA{\textbf{0.0187}}\def\valBADBDA{\textbf{0.0452}}
\def\valBADAAB{0.0781}\def\valBADABB{0.0490}\def\valBADACB{\textbf{0.0182}}\def\valBADADB{0.0484}\def\valBADBAB{0.0703}\def\valBADBBB{0.0377}\def\valBADBCB{0.0175}\def\valBADBDB{0.0418}
\def\valBADAAC{0.0612}\def\valBADABC{0.0396}\def\valBADACC{0.0170}\def\valBADADC{0.0393}\def\valBADBAC{0.0683}\def\valBADBBC{0.0371}\def\valBADBCC{0.0143}\def\valBADBDC{0.0399}
\def\valBADAAD{0.0775}\def\valBADABD{0.0426}\def\valBADACD{0.0147}\def\valBADADD{0.0449}\def\valBADBAD{0.0675}\def\valBADBBD{0.0379}\def\valBADBCD{0.0143}\def\valBADBDD{0.0399}
\def\valBBAAAA{0.2174}\def\valBBAABA{0.3657}\def\valBBAACA{0.4596}\def\valBBAADA{0.3476}\def\valBBABAA{0.1348}\def\valBBABBA{0.2265}\def\valBBABCA{0.2965}\def\valBBABDA{0.2193}
\def\valBBAAAB{\textbf{0.2337}}\def\valBBAABB{0.4345}\def\valBBAACB{0.6021}\def\valBBAADB{0.4234}\def\valBBABAB{0.1422}\def\valBBABBB{0.2601}\def\valBBABCB{0.3494}\def\valBBABDB{0.2506}
\def\valBBAAAD{0.2264}\def\valBBAABD{\textbf{0.4782}}\def\valBBAACD{\textbf{0.6095}}\def\valBBAADD{\textbf{0.4380}}\def\valBBABAD{\textbf{0.2124}}\def\valBBABBD{\textbf{0.2859}}\def\valBBABCD{\textbf{0.4021}}\def\valBBABDD{\textbf{0.3001}}
\def\valBBBAAA{0.2112}\def\valBBBABA{0.3723}\def\valBBBACA{0.4882}\def\valBBBADA{0.3572}\def\valBBBBAA{0.1325}\def\valBBBBBA{0.2248}\def\valBBBBCA{0.2951}\def\valBBBBDA{0.2175}
\def\valBBBAAB{0.2214}\def\valBBBABB{0.4259}\def\valBBBACB{0.5400}\def\valBBBADB{0.3958}\def\valBBBBAB{0.1400}\def\valBBBBBB{0.2584}\def\valBBBBCB{0.3481}\def\valBBBBDB{0.2488}
\def\valBBBAAD{\textbf{0.3031}}\def\valBBBABD{\textbf{0.4685}}\def\valBBBACD{\textbf{0.6110}}\def\valBBBADD{\textbf{0.4609}}\def\valBBBBAD{\textbf{0.1503}}\def\valBBBBBD{\textbf{0.3059}}\def\valBBBBCD{\textbf{0.3880}}\def\valBBBBDD{\textbf{0.2814}}
\def\valBBCAAA{0.0898}\def\valBBCABA{0.1644}\def\valBBCACA{0.2293}\def\valBBCADA{0.1612}\def\valBBCBAA{0.0548}\def\valBBCBBA{0.0995}\def\valBBCBCA{0.1451}\def\valBBCBDA{0.0998}
\def\valBBCAAB{0.0978}\def\valBBCABB{0.1728}\def\valBBCACB{0.2353}\def\valBBCADB{0.1686}\def\valBBCBAB{0.0597}\def\valBBCBBB{0.1046}\def\valBBCBCB{0.1489}\def\valBBCBDB{0.1044}
\def\valBBCAAC{\textbf{0.1408}}\def\valBBCABC{\textbf{0.2073}}\def\valBBCACC{\textbf{0.2488}}\def\valBBCADC{\textbf{0.1990}}\def\valBBCBAC{\textbf{0.1007}}\def\valBBCBBC{\textbf{0.1384}}\def\valBBCBCC{\textbf{0.1514}}\def\valBBCBDC{\textbf{0.1302}}
\def\valBBCAAD{0.1003}\def\valBBCABD{0.1745}\def\valBBCACD{0.2391}\def\valBBCADD{0.1713}\def\valBBCBAD{0.0612}\def\valBBCBBD{0.1056}\def\valBBCBCD{0.1513}\def\valBBCBDD{0.1060}
\def\valBBDAAA{\textbf{0.0809}}\def\valBBDABA{0.0414}\def\valBBDACA{0.0168}\def\valBBDADA{0.0464}\def\valBBDBAA{0.0710}\def\valBBDBBA{0.0411}\def\valBBDBCA{0.0163}\def\valBBDBDA{0.0428}
\def\valBBDAAB{0.0785}\def\valBBDABB{\textbf{0.0487}}\def\valBBDACB{0.0164}\def\valBBDADB{\textbf{0.0479}}\def\valBBDBAB{0.0633}\def\valBBDBBB{\textbf{0.0420}}\def\valBBDBCB{0.0156}\def\valBBDBDB{0.0403}
\def\valBBDAAC{0.0657}\def\valBBDABC{0.0397}\def\valBBDACC{0.0157}\def\valBBDADC{0.0404}\def\valBBDBAC{\textbf{0.0755}}\def\valBBDBBC{0.0410}\def\valBBDBCC{0.0152}\def\valBBDBDC{\textbf{0.0439}}
\def\valBBDAAD{0.0792}\def\valBBDABD{0.0420}\def\valBBDACD{\textbf{0.0186}}\def\valBBDADD{0.0466}\def\valBBDBAD{0.0604}\def\valBBDBBD{0.0394}\def\valBBDBCD{\textbf{0.0167}}\def\valBBDBDD{0.0388}

\centering
\caption{
The results of task 2 with different settings.
}  
\resizebox{1.0\textwidth}{!}{
\begin{tabular}{llcccccccc}  
\toprule[2pt]
\multirow{2}{*}{\hspace{1mm}\textbf{Type}}	&		\multirow{2}{*}{\hspace{2mm}\textbf{Model}}	&		\multicolumn{4}{c}{\textbf{Simple Questions}}										&		\multicolumn{4}{c}{\textbf{Complex Questions}}										\\
						\cmidrule(lr){3-6}												\cmidrule(lr){7-10}										
	&			&		\textbf{Top-1}	&		\textbf{Top-10}	&		\textbf{Top-50}	&		\textbf{Avg.}	&		\textbf{Top-1}	&		\textbf{Top-10}	&		\textbf{Top-5}	&		\textbf{Avg.}	\\
\midrule
\multicolumn{10}{c}{\textbf{Chinese Questions}}	\\
\midrule
\cellcolor{mycolor} \textbf{Vid. + Sub. + K.G.}	&	\cellcolor{mycolor}	$\begin{array}{l} \text{CCGS \cite{CCGS}} \\ \text{FMALG \cite{FMALG}} \\ \text{Qwen-2.5-VL \cite{Qwen}} \end{array}$	&	\cellcolor{mycolor}	$\begin{array}{c} \valBAAAAA \\ \valBAAAAB \\ \valBAAAAD \end{array}$	&	\cellcolor{mycolor}	$\begin{array}{c} \valBAAABA \\ \valBAAABB \\ \valBAAABD \end{array}$	&	\cellcolor{mycolor}	$\begin{array}{c} \valBAAACA \\ \valBAAACB \\ \valBAAACD \end{array}$	&	\cellcolor{mycolor}	$\begin{array}{c} \valBAAADA \\ \valBAAADB \\ \valBAAADD \end{array}$	&	\cellcolor{mycolor}	$\begin{array}{c} \valBAABAA \\ \valBAABAB \\ \valBAABAD \end{array}$	&	\cellcolor{mycolor}	$\begin{array}{c} \valBAABBA \\ \valBAABBB \\ \valBAABBD \end{array}$	&	\cellcolor{mycolor}	$\begin{array}{c} \valBAABCA \\ \valBAABCB \\ \valBAABCD \end{array}$	&	\cellcolor{mycolor}	$\begin{array}{c} \valBAABDA \\ \valBAABDB \\ \valBAABDD \end{array}$	\\
\cellcolor{baise} \textbf{Vid. + Sub.}	&	\cellcolor{baise}	$\begin{array}{l} \text{CCGS \cite{CCGS}} \\ \text{FMALG \cite{FMALG}} \\ \text{Qwen-2.5-VL \cite{Qwen}} \end{array}$	&	\cellcolor{baise}	$\begin{array}{c} \valBABAAA \\ \valBABAAB \\ \valBABAAD \end{array}$	&	\cellcolor{baise}	$\begin{array}{c} \valBABABA \\ \valBABABB \\ \valBABABD \end{array}$	&	\cellcolor{baise}	$\begin{array}{c} \valBABACA \\ \valBABACB \\ \valBABACD \end{array}$	&	\cellcolor{baise}	$\begin{array}{c} \valBABADA \\ \valBABADB \\ \valBABADD \end{array}$	&	\cellcolor{baise}	$\begin{array}{c} \valBABBAA \\ \valBABBAB \\ \valBABBAD \end{array}$	&	\cellcolor{baise}	$\begin{array}{c} \valBABBBA \\ \valBABBBB \\ \valBABBBD \end{array}$	&	\cellcolor{baise}	$\begin{array}{c} \valBABBCA \\ \valBABBCB \\ \valBABBCD \end{array}$	&	\cellcolor{baise}	$\begin{array}{c} \valBABBDA \\ \valBABBDB \\ \valBABBDD \end{array}$	\\
\cellcolor{mycolor} \textbf{Sub. + K.G.}	&	\cellcolor{mycolor}	$\begin{array}{l} \text{CCGS \cite{CCGS}} \\ \text{FMALG \cite{FMALG}} \\ \text{GPT-4o \cite{GPT}} \\ \text{Qwen-2.5-VL \cite{Qwen}} \end{array}$	&	\cellcolor{mycolor}	$\begin{array}{c} \valBACAAA \\ \valBACAAB \\ \valBACAAC \\ \valBACAAD \end{array}$	&	\cellcolor{mycolor}	$\begin{array}{c} \valBACABA \\ \valBACABB \\ \valBACABC \\ \valBACABD \end{array}$	&	\cellcolor{mycolor}	$\begin{array}{c} \valBACACA \\ \valBACACB \\ \valBACACC \\ \valBACACD \end{array}$	&	\cellcolor{mycolor}	$\begin{array}{c} \valBACADA \\ \valBACADB \\ \valBACADC \\ \valBACADD \end{array}$	&	\cellcolor{mycolor}	$\begin{array}{c} \valBACBAA \\ \valBACBAB \\ \valBACBAC \\ \valBACBAD \end{array}$	&	\cellcolor{mycolor}	$\begin{array}{c} \valBACBBA \\ \valBACBBB \\ \valBACBBC \\ \valBACBBD \end{array}$	&	\cellcolor{mycolor}	$\begin{array}{c} \valBACBCA \\ \valBACBCB \\ \valBACBCC \\ \valBACBCD \end{array}$	&	\cellcolor{mycolor}	$\begin{array}{c} \valBACBDA \\ \valBACBDB \\ \valBACBDC \\ \valBACBDD \end{array}$	\\
\cellcolor{baise} \textbf{Sub.}	&	\cellcolor{baise}	$\begin{array}{l} \text{CCGS \cite{CCGS}} \\ \text{FMALG \cite{FMALG}} \\ \text{GPT-4o \cite{GPT}} \\ \text{Qwen-2.5-VL \cite{Qwen}} \end{array}$	&	\cellcolor{baise}	$\begin{array}{c} \valBADAAA \\ \valBADAAB \\ \valBADAAC \\ \valBADAAD \end{array}$	&	\cellcolor{baise}	$\begin{array}{c} \valBADABA \\ \valBADABB \\ \valBADABC \\ \valBADABD \end{array}$	&	\cellcolor{baise}	$\begin{array}{c} \valBADACA \\ \valBADACB \\ \valBADACC \\ \valBADACD \end{array}$	&	\cellcolor{baise}	$\begin{array}{c} \valBADADA \\ \valBADADB \\ \valBADADC \\ \valBADADD \end{array}$	&	\cellcolor{baise}	$\begin{array}{c} \valBADBAA \\ \valBADBAB \\ \valBADBAC \\ \valBADBAD \end{array}$	&	\cellcolor{baise}	$\begin{array}{c} \valBADBBA \\ \valBADBBB \\ \valBADBBC \\ \valBADBBD \end{array}$	&	\cellcolor{baise}	$\begin{array}{c} \valBADBCA \\ \valBADBCB \\ \valBADBCC \\ \valBADBCD \end{array}$	&	\cellcolor{baise}	$\begin{array}{c} \valBADBDA \\ \valBADBDB \\ \valBADBDC \\ \valBADBDD \end{array}$	\\
\cellcolor{mycolor} \textbf{Human} & \cellcolor{mycolor} \hspace{1cm} \textbf{---}  & \cellcolor{mycolor} $\begin{array}{c} 0.7133 \end{array}$ & \cellcolor{mycolor} $\begin{array}{c} 0.7471 \end{array}$ & \cellcolor{mycolor} $\begin{array}{c} 0.7983 \end{array}$ & \cellcolor{mycolor} $\begin{array}{c} 0.7529 \end{array}$ & \cellcolor{mycolor} $\begin{array}{c} 0.7092 \end{array}$ & \cellcolor{mycolor} $\begin{array}{c} 0.7360 \end{array}$ & \cellcolor{mycolor} $\begin{array}{c} 0.7876 \end{array}$ & \cellcolor{mycolor} $\begin{array}{c} 0.7443 \end{array}$ \\
\midrule
\multicolumn{10}{c}{\textbf{English Questions}}	\\
\midrule
\cellcolor{mycolor} \textbf{Vid. + Sub. + K.G.}	&	\cellcolor{mycolor}	$\begin{array}{l} \text{CCGS \cite{CCGS}} \\ \text{FMALG \cite{FMALG}} \\ \text{Qwen-2.5-VL \cite{Qwen}} \end{array}$	&	\cellcolor{mycolor}	$\begin{array}{c} \valBBAAAA \\ \valBBAAAB \\ \valBBAAAD \end{array}$	&	\cellcolor{mycolor}	$\begin{array}{c} \valBBAABA \\ \valBBAABB \\ \valBBAABD \end{array}$	&	\cellcolor{mycolor}	$\begin{array}{c} \valBBAACA \\ \valBBAACB \\ \valBBAACD \end{array}$	&	\cellcolor{mycolor}	$\begin{array}{c} \valBBAADA \\ \valBBAADB \\ \valBBAADD \end{array}$	&	\cellcolor{mycolor}	$\begin{array}{c} \valBBABAA \\ \valBBABAB \\ \valBBABAD \end{array}$	&	\cellcolor{mycolor}	$\begin{array}{c} \valBBABBA \\ \valBBABBB \\ \valBBABBD \end{array}$	&	\cellcolor{mycolor}	$\begin{array}{c} \valBBABCA \\ \valBBABCB \\ \valBBABCD \end{array}$	&	\cellcolor{mycolor}	$\begin{array}{c} \valBBABDA \\ \valBBABDB \\ \valBBABDD \end{array}$	\\
\cellcolor{baise} \textbf{Vid. + Sub.}	&	\cellcolor{baise}	$\begin{array}{l} \text{CCGS \cite{CCGS}} \\ \text{FMALG \cite{FMALG}} \\ \text{Qwen-2.5-VL \cite{Qwen}} \end{array}$	&	\cellcolor{baise}	$\begin{array}{c} \valBBBAAA \\ \valBBBAAB \\ \valBBBAAD \end{array}$	&	\cellcolor{baise}	$\begin{array}{c} \valBBBABA \\ \valBBBABB \\ \valBBBABD \end{array}$	&	\cellcolor{baise}	$\begin{array}{c} \valBBBACA \\ \valBBBACB \\ \valBBBACD \end{array}$	&	\cellcolor{baise}	$\begin{array}{c} \valBBBADA \\ \valBBBADB \\ \valBBBADD \end{array}$	&	\cellcolor{baise}	$\begin{array}{c} \valBBBBAA \\ \valBBBBAB \\ \valBBBBAD \end{array}$	&	\cellcolor{baise}	$\begin{array}{c} \valBBBBBA \\ \valBBBBBB \\ \valBBBBBD \end{array}$	&	\cellcolor{baise}	$\begin{array}{c} \valBBBBCA \\ \valBBBBCB \\ \valBBBBCD \end{array}$	&	\cellcolor{baise}	$\begin{array}{c} \valBBBBDA \\ \valBBBBDB \\ \valBBBBDD \end{array}$	\\
\cellcolor{mycolor} \textbf{Sub. + K.G.}	&	\cellcolor{mycolor}	$\begin{array}{l} \text{CCGS \cite{CCGS}} \\ \text{FMALG \cite{FMALG}} \\ \text{GPT-4o \cite{GPT}} \\ \text{Qwen-2.5-VL \cite{Qwen}} \end{array}$	&	\cellcolor{mycolor}	$\begin{array}{c} \valBBCAAA \\ \valBBCAAB \\ \valBBCAAC \\ \valBBCAAD \end{array}$	&	\cellcolor{mycolor}	$\begin{array}{c} \valBBCABA \\ \valBBCABB \\ \valBBCABC \\ \valBBCABD \end{array}$	&	\cellcolor{mycolor}	$\begin{array}{c} \valBBCACA \\ \valBBCACB \\ \valBBCACC \\ \valBBCACD \end{array}$	&	\cellcolor{mycolor}	$\begin{array}{c} \valBBCADA \\ \valBBCADB \\ \valBBCADC \\ \valBBCADD \end{array}$	&	\cellcolor{mycolor}	$\begin{array}{c} \valBBCBAA \\ \valBBCBAB \\ \valBBCBAC \\ \valBBCBAD \end{array}$	&	\cellcolor{mycolor}	$\begin{array}{c} \valBBCBBA \\ \valBBCBBB \\ \valBBCBBC \\ \valBBCBBD \end{array}$	&	\cellcolor{mycolor}	$\begin{array}{c} \valBBCBCA \\ \valBBCBCB \\ \valBBCBCC \\ \valBBCBCD \end{array}$	&	\cellcolor{mycolor}	$\begin{array}{c} \valBBCBDA \\ \valBBCBDB \\ \valBBCBDC \\ \valBBCBDD \end{array}$	\\
\cellcolor{baise} \textbf{Sub.}	&	\cellcolor{baise}	$\begin{array}{l} \text{CCGS \cite{CCGS}} \\ \text{FMALG \cite{FMALG}} \\ \text{GPT-4o \cite{GPT}} \\ \text{Qwen-2.5-VL \cite{Qwen}} \end{array}$	&	\cellcolor{baise}	$\begin{array}{c} \valBBDAAA \\ \valBBDAAB \\ \valBBDAAC \\ \valBBDAAD \end{array}$	&	\cellcolor{baise}	$\begin{array}{c} \valBBDABA \\ \valBBDABB \\ \valBBDABC \\ \valBBDABD \end{array}$	&	\cellcolor{baise}	$\begin{array}{c} \valBBDACA \\ \valBBDACB \\ \valBBDACC \\ \valBBDACD \end{array}$	&	\cellcolor{baise}	$\begin{array}{c} \valBBDADA \\ \valBBDADB \\ \valBBDADC \\ \valBBDADD \end{array}$	&	\cellcolor{baise}	$\begin{array}{c} \valBBDBAA \\ \valBBDBAB \\ \valBBDBAC \\ \valBBDBAD \end{array}$	&	\cellcolor{baise}	$\begin{array}{c} \valBBDBBA \\ \valBBDBBB \\ \valBBDBBC \\ \valBBDBBD \end{array}$	&	\cellcolor{baise}	$\begin{array}{c} \valBBDBCA \\ \valBBDBCB \\ \valBBDBCC \\ \valBBDBCD \end{array}$	&	\cellcolor{baise}	$\begin{array}{c} \valBBDBDA \\ \valBBDBDB \\ \valBBDBDC \\ \valBBDBDD \end{array}$	\\
\cellcolor{mycolor} \textbf{Human} & \cellcolor{mycolor} \hspace{1cm} \textbf{---}  & \cellcolor{mycolor} $\begin{array}{c} 0.7039 \end{array}$ & \cellcolor{mycolor} $\begin{array}{c} 0.7260 \end{array}$ & \cellcolor{mycolor} $\begin{array}{c} 0.7434 \end{array}$ & \cellcolor{mycolor} $\begin{array}{c} 0.7244 \end{array}$ & \cellcolor{mycolor} $\begin{array}{c} 0.7077 \end{array}$ & \cellcolor{mycolor} $\begin{array}{c} 0.7225 \end{array}$ & \cellcolor{mycolor} $\begin{array}{c} 0.7429 \end{array}$ & \cellcolor{mycolor} $\begin{array}{c} 0.7243 \end{array}$ \\
            \bottomrule[2pt]  
        \end{tabular}    
    }
    \label{result2}  
\begin{tablenotes}
\footnotesize
\item[*] \textbf{Bold} represents the grade of the top-performing model in the current setting.
Top-1: (Top-1 IoU score);
Top-10: (Top-10 IoU score);
Top-50: (Top-50 IoU score);
mIoU: (The mean IoU of all sample);
Type: (The type of settings when challengers solving the problem);
Vid.: (Videos);
Sub.: (Subtitles in text format);
K.G.: (Knowledge graphs in text format);
Human: (Tasks solved manually).
\end{tablenotes}
\end{table}

\subsection{Experiments Settings}
We selected several state-of-the-art (SOTA) models as our primary baselines. For the TAGSV task, we evaluated MutualSL \cite{MutualSL} and PBL \cite{PBL}. For the TAGVC task, we evaluated CCGS \cite{CCGS} and FMALG \cite{FMALG}.

In addition to these specialized models, we also evaluated the zero-shot performance of two powerful, general-purpose text-only LLMs and MLLMs: GPT-4o \cite{GPT} and Qwen2.5-VL \cite{QwenVL}. These (M)LLMs were tested without any fine-tuning, guided only by carefully designed system prompts.

To precisely measure the contribution of each modality and validate the necessity of multi-modal, multi-hop reasoning, we designed a series of ablation studies with different input settings:
(1) \textbf{Full Input}: Models receive subtitles, video frames, and the knowledge graph.
(2) \textbf{Video + Subtitles}: Models receive both subtitles and video frames.
(3) \textbf{Subtitles + KG}: Models receive subtitles and the knowledge graph.
(4) \textbf{Subtitles Only}: Models receive only the subtitles.

The Subtitles + KG setting requires careful consideration. Since our ground-truth KGs were manually created using visual information, providing them as plain text could create a ``shortcut," potentially allowing models to bypass true visual reasoning. 
This setting is therefore included specifically to quantify the performance gain from structured knowledge, while acknowledging this potential information leak.

Finally, to establish an upper bound and confirm that the tasks are solvable, we let human challengers who were not involved in the annotation process to complete the same tasks.

\subsection{Results and Analysis}
The comprehensive results for both the TAGSV and TAGVC tasks are presented in Tabs. \ref{result1} and \ref{result2}, respectively.

\textbf{Overall Performance Trends.}
Across all models and tasks, several clear trends emerge from the results. 
First, performance on simple questions is consistently higher than on complex questions, confirming that our complex questions successfully introduce a greater reasoning challenge. 
Second, as expected, model performance improves as more modalities are provided, with the Full Input setting yielding the best results, followed by Video + Subtitles, Subtitles + KG and finally the Subtitles Only baseline. 
This directly validates the multi-modal nature of our benchmark.

\textbf{Impact of the Knowledge Graph.}
The inclusion of the ground-truth Knowledge Graph (Subtitles + KG) provides a significant performance boost over the Subtitles Only setting. 
This demonstrates that structured, external knowledge is crucial for resolving complex reasoning chains, even when it is provided as plain text. 
The KG helps models understand entities and relationships not explicitly stated in the subtitles, allowing for more nuanced judgments.

\textbf{Comparison of Model Architectures.}
A key finding is that the general-purpose (M)LLMs (GPT-4o, Qwen2.5-VL), despite being tested in a zero-shot setting, often outperform the specialized VTG models that were designed for similar tasks. 
This highlights the powerful emergent reasoning capabilities of modern large-scale models.

\textbf{Language Consistency and Gap to Human Performance.}
We observe that most models achieve comparable performance on both the Chinese and English subsets, likely due to the extensive multilingual data used in their pre-training. 
However, even the best-performing model still lags significantly behind the human challengers baseline. This substantial gap underscores the core challenge presented by M$^3$-Med and indicates that there is ample room for improvement in deep semantic understanding, complex reasoning, and precise temporal localization in videos.

\subsection{Simple Questions and Complex Questions}
While the performance gap in our main experiments suggests a clear distinction between simple and complex questions, we conducted a deeper linguistic analysis to provide further quantitative evidence for this divide.

\textbf{Lexical Similarity.} 
We first measured the lexical overlap between the questions and the subtitles of their corresponding answer segments by calculating the IoU of their word sets after lemmatization for English text and tokenization for Chinese text.

As shown in Fig. \ref{Analysis1}, the lexical overlap for our simple questions is comparable to that of existing benchmarks like MedVidQA and HealthVidQA. In stark contrast, the overlap for our complex questions is significantly lower, confirming they are intentionally designed to prevent simple keyword matching.

\textbf{Semantic Relevance.}
To ensure that complex questions still pertain to the content, we also computed semantic similarity scores. We used a pre-trained S-BERT model \cite{SBERT} to measure relevance between the question and subtitles (Fig. \ref{Analysis2}) and the multilingual AltCLIP model \cite{AltCLIP} to measure relevance between the question and the corresponding video frames (Fig. \ref{Analysis3}).

\begin{figure}[ht!] 
    \input{Figures/6.1.Relevance}
\end{figure}

\textbf{Summary.} This analysis confirms our design goal: M$^3$-Med's complex questions feature low lexical overlap but high semantic and visual relevance. This combination creates a significant challenge for models that rely on simple text-matching shortcuts and forces them to engage in deeper, cross-modal reasoning.
            \section{Discussion}

\subsection{The Nuanced Role of Explicit Knowledge}
Our results present a nuanced view of the role of structured knowledge. Contrary to our initial hypothesis that a ground-truth Knowledge Graph might serve as a powerful ``shortcut," the performance gains were moderate (Tabs. \ref{result1} and \ref{result2}). 
This suggests that while explicit knowledge is beneficial, it cannot fully replace genuine visual reasoning. 
Models must still perform a difficult synthesis of symbolic (KG) and perceptual (video) information, which remains a core challenge.

\subsection{The Challenge of (M)LLMs: Beyond Accuracy}
While (M)LLMs demonstrate superior performance in terms of accuracy, our qualitative analysis reveals unique failure modes not seen in specialized models. A primary issue is output format adherence; despite simple prompting, the (M)LLMs frequently produced malformed outputs (see Tab. \ref{Wrong}).

\begin{table}[h]

    \centering  
    \caption{Qualitative Examples of Malformed Outputs from (multi-modal) large language models. This table showcases typical formatting errors and non-standard outputs generated by baseline (M)LLMs, underscoring the challenges these models face in adhering to the required answer format.}  
    \resizebox{1.0\textwidth}{!}{
        \begin{tabular}{ll}  
            \toprule[2pt] 
\textbf{Description}		&	\textbf{Example} \\
\midrule												
\rowcolor{mycolor} Wrong time format(hh:mm:ss or mm:ss).	& 00:01:15 00:01:35 \\
\rowcolor{baise} Only one time-stamp is given.	& UyJRSVhZfIU 58 \\
\rowcolor{mycolor} The number of seconds that are not an integer.	& 15.6 23.2 \\
\rowcolor{baise} Extra connectors.	& 00:02:18$\sim$00:03:15 \\
\rowcolor{mycolor} Extra quotes.	& ``25 34" \\
\rowcolor{baise} Extra text.	& Okay, according to your requirements, find out... \\
            \bottomrule[2pt]  
        \end{tabular}    
    }
    \label{Wrong}  
\end{table}

Furthermore, we observed a phenomenon of spurious refusals, where models would output ``Null" or ``Cannot answer" even when a correct answer existed. 
This contrasts with traditional fine-tuned models, which are heavily penalized for refusal and thus rarely do so. 
This may indicate that general-purpose (M)LLMs have different internal mechanisms for handling uncertainty, a behavior that warrants further investigation, as it has significant implications for their reliability in structured, real-world tasks.

            \section{Limitations}
Our work, while comprehensive, has several limitations that should be acknowledged.
\subsection{Video Sources and Ethical Considerations}
The videos in our dataset are sourced from public platforms like YouTube, which, while providing diverse content, introduces potential copyright issues for certain downstream applications.

We must also acknowledge two potential sources of error. 
First, while manually screened, the source videos themselves may contain factual inaccuracies. Second, any model trained on this data is still susceptible to hallucination, a prevalent issue in the AI field. 
Therefore, we strongly recommend that any system developed using this benchmark, especially in sensitive medical contexts, should include a clear disclaimer stating that its outputs are for informational purposes only and not a substitute for professional medical advice.

\subsection{Annotation Scalability and Automation}
To manage the significant cost of annotation, we aimed to automate the pipeline wherever possible. For instance, the use of Whisper \cite{Whisper} for subtitle generation proved highly effective and scalable.
However, fully automating the knowledge graph construction, particularly the relation extraction step, remains a challenge. 
Our experiments with existing cross-modal models yielded unsatisfactory results, necessitating significant human-in-the-loop curation. 
Consequently, our pipeline is currently semi-automated, and achieving full, high-quality automation for such complex annotations is a key limitation and an important direction for future research.
            \section{Conclusion and Future Work}
In this paper, we presented M$^3$-Med, a novel benchmark designed to push the boundaries of video understanding towards multi-lingual, multi-modal, and multi-hop reasoning. 
Through a rigorous, expert-guided annotation process, we developed two structured tasks, TAGSV and TAGVC, that effectively distinguish between shallow pattern matching and deep, causal reasoning.

Our comprehensive experiments reveal a significant performance gap between current SOTA models and human experts, particularly on complex questions requiring genuine cross-modal synthesis. 
Even powerful (M)LLMs, while promising, struggle to coherently integrate visual and textual evidence and exhibit unique failure modes, highlighting critical areas for improvement.

Ultimately, M$^3$-Med serves not just as a static benchmark, but as a dynamic challenge to the research community: to build the next generation of models capable of robust, explainable, and reliable reasoning in high-stakes domains like medicine.

\textbf{Future Work.} 
While our work emphasizes the quality of manual annotation, the cost and scalability limitations motivate our primary future goal: exploring novel semi-automated annotation pipelines that can reduce costs while preserving the high-quality, multi-hop nature of our benchmark.

Beyond scaling our dataset, future research avenues include: 
(1) Model Development: Investigating domain-specific fine-tuning strategies for (M)LLMs to improve their reliability and instruction-following. 
(2) Task Expansion: Extending the benchmark to support more interactive tasks, such as conversational QA. 
(3) Ethical Frameworks: Developing robust protocols for model verification and deployment in safety-critical medical scenarios.
            
		\Acknowledgements{This work was supported by National Natural Science Foundation of China (Nos. 4222037 and L181010), and Sanming Project of Medicine in Shenzhen (No. SZZYSM202311002).}
		
		

        \bibliographystyle{unsrt}
        \bibliography{reference}

			
\end{document}